\documentclass{article}

 \usepackage[preprint]{open_neurips}

\usepackage[utf8]{inputenc} % allow utf-8 input
\usepackage[T1]{fontenc}    % use 8-bit T1 fonts
\usepackage{hyperref}       %
\hypersetup{
    colorlinks=true,
    linkcolor=bigred,
    citecolor=blue,
    filecolor=green,      
    urlcolor=darkred,
    }
\usepackage{url}            % simple URL typesetting
\usepackage{booktabs}       % professional-quality tables
\usepackage{amsmath}
\usepackage{amssymb}
\usepackage{amsfonts}       % blackboard math symbols
\usepackage{nicefrac}       % compact symbols for 1/2, etc.
\usepackage{microtype}      % microtypography
\usepackage{xcolor}         % colors
\usepackage[dvipsnames]{xcolor}
\usepackage{listings}
\usepackage{svg}
\usepackage{multirow}
\usepackage[table]{xcolor}  % for \rowcolor
\usepackage{array}          % safer table formatting
\usepackage{subcaption}
\lstdefinestyle{paperstyle}{
    basicstyle=\ttfamily\small,
    breaklines=true,
    columns=fullflexible,
    frame=single,
    rulecolor=\color{gray!30},
    backgroundcolor=\color{gray!5},
    captionpos=b,
    xleftmargin=1.5em,
    xrightmargin=1.5em,
    aboveskip=0.5em,
    belowskip=0.5em,
    literate={␣}{{\ }}1,
}

\usepackage{tikz}

\newlength\savewidth\newcommand\shline{\noalign{\global\savewidth\arrayrulewidth
  \global\arrayrulewidth 1pt}\hline\noalign{\global\arrayrulewidth\savewidth}}
\newcolumntype{x}[1]{>{\centering\arraybackslash}p{#1pt}}
\newcolumntype{y}[1]{>{\raggedright\arraybackslash}p{#1pt}}
\newcolumntype{z}[1]{>{\raggedleft\arraybackslash}p{#1pt}}
\definecolor{mygreen}{RGB}{0, 205, 108}
\definecolor{convcolor}{HTML}{412F8A}
\definecolor{resnetcolor}{HTML}{8DA0CB}
\definecolor{vitcolor}{HTML}{fc8e62}

\newcommand{\tablestyle}[2]{\setlength{\tabcolsep}{#1}\renewcommand{\arraystretch}{#2}\centering\footnotesize}

\definecolor{orange}{HTML}{ff7f0e}
\definecolor{darkred}{HTML}{c00000}     %
\definecolor{bigred}{HTML}{8a0000}     %
\definecolor{darkgreen}{HTML}{66CC66}   %
\definecolor{baselinecolor}{gray}{.9}

\definecolor{lightyellow}{rgb}{1.0, 0.98, 0.8}
\definecolor{lightblue}{rgb}{0.85, 0.9, 1.0}
\definecolor{lightgreen}{rgb}{0.85, 1.0, 0.9}
\definecolor{nvidiagreen}{rgb}{0.88, 0.96, 0.82}
\definecolor{lightorange}{rgb}{0.99, 0.83, 0.64}
\definecolor{baselinecolor}{rgb}{0.9, 0.9, 0.9}

\title{COPRA: Conditional Parameter Adaptation with Reinforcement Learning for Video Anomaly Detection}

\author{%
  Darryl Cherian Jacob$^{1}$ \quad Xinyu Liu$^{1}$ \quad Kai Wang$^{2}$ \quad Pan He$^{1}$\textsuperscript{\textdagger}  \\ 
  $^{1}$Auburn University \quad $^{2}$Tencent Hunyuan \\  \textsuperscript{\textdagger}Corresponding author
}

\begin{document}

\maketitle

% \begingroup
% \renewcommand\thefootnote{\textdagger}
% \footnotetext{Corresponding author.}
% \endgroup

\begin{abstract}
 
Vision-language models (VLMs) have demonstrated impressive performance in video anomaly detection (VAD), while providing interpretable predictions. However, we identify a key yet underexplored VAD issue: the mismatch between training and inference in both data distribution and model configuration. First, many VLM-based VAD approaches rely on a static parameter paradigm after training (or post-adaptation), causing them to overfit the training distribution and limiting generalization to unseen domains. 
Consequently, these methods lack effective test-time adaptation under substantial distribution shifts, such as novel environments or previously unseen anomaly types. Second, they treat VLMs as reasoning-based classifiers trained on extremely sparse frames from long videos, but perform inference on densely sampled short segments via sliding windows, leading to inherent inconsistencies between training and testing. To address these limitations, we propose COPRA, a conditional parameter adaptation framework for VLM-based VAD. Rather than relying on static post-training adaptation (e.g., fixed prompts or shared parameter updates), COPRA introduces instance-conditioned parameter adaptation, where a lightweight generator predicts input-specific parameter updates to dynamically modulate a frozen VLM on the fly, enabling per-segment adaptation during both training and inference. Empirically, our method achieves strong performance on standard VAD benchmarks.
More broadly, it consistently outperforms static baselines in both in-domain and cross-domain settings, and further generalizes beyond VAD to unseen tasks such as multiple-choice Video Question Answering and Dense Captioning. These results suggest that COPRA provides an effective weight space generation mechanism for foundation models, enabling more scalable, adaptive, and context-aware video understanding. The code will be released at https://github.com/THE-MALT-LAB/COPRA
\end{abstract}

\section{Introduction}

Video anomaly detection (VAD) seeks to identify and localize rare and unexpected events in long videos, often under significant variability in scene context, camera viewpoints, and anomaly types. Recent VAD approaches based on vision-language models (VLMs)~\cite{zanella2024harnessing,zhang2024holmes,ye2025vera,zhu2025vau} are particularly promising, via leveraging rich semantic priors from large-scale pretraining and enabling interpretable predictions. 

\begin{figure}
    \centering
    \includegraphics[width=\linewidth]{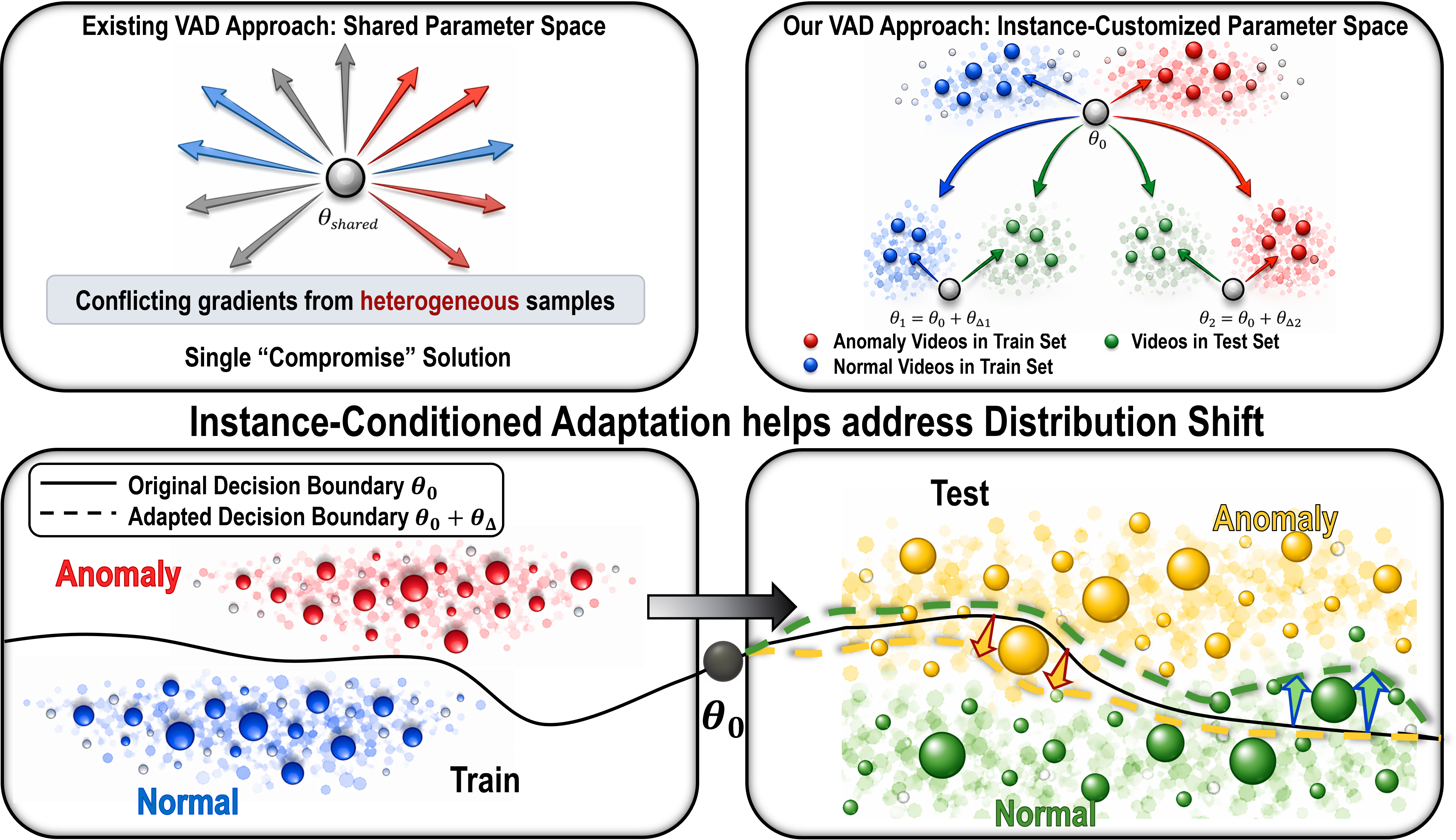}%
    \caption{Illustration of the generalization gap in VAD caused by training--inference mismatch and train--test distribution shift. Existing methods rely on shared, static parameters, forcing diverse video patterns into a single compromise solution. As a result, \textit{the decision boundary learned from the training dataset often fails to generalize to unseen test distributions, causing some samples to fall on the wrong side of the boundary}. We address this limitation through instance-conditioned parameter adaptation, where each video \textit{dynamically pulls the decision landscape toward a more suitable boundary via input-specific parameter updates}, improving robustness and generalization.}
   \label{fig:main_problem_visualization}
\end{figure}

However, effectively adapting VLMs to VAD remains an actively evolving and largely underexplored challenge. As illustrated in Figure~\ref{fig:main_problem_visualization}, existing VAD approaches~\cite{zanella2024harnessing,zhang2024holmes,ye2025vera,huang2025ex,zhu2025vau,zou2026unlocking} typically learn a single, shared parameter space, despite the substantial heterogeneity across anomaly types, temporal dynamics, and scene contexts. During training, each heterogeneous sample induces a distinct gradient direction that may conflict with others, leading to parameters that converge to a compromise rather than being optimal for any single sample~\cite{hywu2026}. Such static, shared adaptations often overfit to dataset-specific patterns encoded in prompts~\cite{ye2025vera} or model parameters~\cite{zhang2024holmes}, rather than capturing transferable notions of abnormality. 

This leads to a pronounced generalization gap, which we term the \textit{shared parameter bottleneck}. For example, detecting a traffic accident~\cite{Kim_2025_ICCV}, physical assault~\cite{sultani2018real}, or suspicious retail behavior requires distinct visual cues, temporal dynamics, and semantic priors. A single shared adaptation must reconcile these heterogeneous demands, leading to degraded robustness and generalization. Moreover, prevailing VAD pipelines exhibit a fundamental training–inference mismatch: models are trained on sparsely sampled frames from long videos with weak video-level supervision, but deployed on densely sampled sliding windows for fine-grained localization. As a result, parameters learned from sparse inputs are applied to dense local decisions, degrading performance under variations in viewpoints, environments, motion patterns, and anomaly semantics.
% leading to significant shifts in model behavior.

In this paper, we advocate an instance-specific perspective on VAD~\cite{hywu2026,han2026survey}. Rather than learning a single shared parameter set, we infer model parameters on the fly for each video segment via instance-conditioned adaptation, where most parameters are shared and frozen, and only a small subset is dynamically customized to emphasize input-specific behaviors. This is motivated by the diversity of anomaly scenarios: some events (e.g., explosions, fires, collisions) can be identified from a few salient frames, while others (e.g., shoplifting, burglary, loitering) require modeling subtle motion patterns and temporal interactions. Consequently, reliable detection spans cues from coarse appearance to fine-grained temporal reasoning, making existing VAD approaches based on a shared-parameter paradigm inherently a compromise. A promising direction is to enable on-the-fly, input-conditioned adaptation that tailors model behavior to each video input. 

% However, existing VAD approaches compress these heterogeneous requirements into a single shared update, forcing the model to reconcile incompatible signals within a globally shared prompt~\cite{ye2025vera} or parameter space~\cite{zhang2024holmes,zhu2025vau}. As a result, the learned adaptation often reflects a compromise that is not well aligned with any specific anomaly regime. We argue that this mismatch is a key source of the generalization challenges in VLM-based VAD. 
Accordingly, we propose COPRA, a conditional parameter adaptation framework for VLM-based VAD that predicts input-specific parameter updates for each video, dynamically modulating a largely frozen VLM on the fly. In practice, COPRA produces instance-conditioned LoRA weights for customized per-input adaptation. As illustrated in Figure~\ref{fig:main_problem_visualization}, COPRA addresses the limitation of existing static adaptation approaches, where \textit{the decision boundary learned from the training dataset often fails to generalize to unseen test distributions}, by dynamically pulling the decision landscape toward a more suitable boundary for each input, thereby improving robustness and generalization under distribution shift. We further employ reinforcement learning fine-tuning to guide token-level parameter generation under video-level supervision and explore flexible frame sampling strategies. 
% Notably, our framework generalizes beyond VAD, demonstrating strong transferability to new tasks.

% \textbf{Contributions}. 
Our main contributions can be summarized as follows:
\begin{enumerate}
\item We identify a fundamental limitation of existing VAD approaches with the \textit{shared parameter bottleneck}, where static, shared adaptations force heterogeneous requirements into a single compromised solution, and advocate a shift toward on-the-fly, input-conditioned adaptation.
\item We present COPRA, the first framework to formulate VAD as an instance-conditioned adaptation problem, where a VLM is dynamically specialized for each video segment.%, independent of the training–testing split.
\item We introduce a conditional parameter generation framework for VLMs that enables dynamic, per-input adaptation through generated LoRA weights. We further incorporate reinforcement learning fine-tuning under video-level supervision and investigate scenario-aware frame sampling strategies. Our framework alleviates training--inference mismatch and reduces cross-domain interference, improving robustness across diverse anomaly patterns.

\item We demonstrate strong performance and robust generalization across standard VAD benchmarks, consistently outperforming static adaptation baselines in both in-domain and cross-dataset settings. Moreover, the learned adaptation transfers effectively beyond VAD to downstream video understanding tasks.
\end{enumerate}

\section{Related Work}

\textbf{VLMs for VAD}. %Parameter-efficient VLM adaptation methods for VAD consistently converge on three architectural choices.   
Recent works leverage VLMs for their strong cross-modal reasoning capabilities~\cite{pratt2023does, liu2023visual}. Existing approaches largely treat VLMs as fixed operators: some attach task-specific detection heads to frozen backbones~\cite{cai2025hiprobe}, others decouple captioning and reasoning via VLM–LLM pipelines~\cite{zanella2024harnessing, yang2024follow}, while prompt-based methods (e.g., VERA~\cite{ye2025vera}) or post-hoc frame selection strategies~\cite{hivau70k, zhang2024holmes} avoid updating model parameters.

% to produce the most information for the reasoner.% Third, recent fine-tuned VLM methods like \cite{violencevlm} are optimized and evaluated on video-level classification.
%Together, these recurring design choices reflect a broader disconnect between current VLM fine-tuning pipelines and frame-level classification in weakly supervised VAD.

\textbf{Shared Parameter Bottleneck in VLM-based VAD.} 
%A deeper structural issue underlies these pitfalls. Whether through full fine-tuning, LoRA \cite{hu2022lora}, or instruction tuning, existing VLM adaptation pipelines for VAD commit to a single shared parameter update that is applied uniformly across all inference inputs. 
As previously observed, existing VLM-based VAD approaches adapt models through shared parameters, either by optimizing global prompts~\cite{ye2025vera} or fine-tuning model weights~\cite{zhang2024holmes}. This introduces a fundamental limitation: model behavior is governed by training data statistics and a fixed operator, hindering generalization to novel videos. As a result, performance degrades on unseen data. Consistent with model editing studies~\cite{hywu2026, han2026survey}, a single shared update is inherently driven toward a compromise, limiting its ability to capture diverse anomaly patterns.

%that partially satisfies none. 

%While they can be effective, they do not explicitly address the underlying training--inference mismatch in VAD, where models are optimized on sparse video-level supervision but deployed under dense segment-level sliding-window inference.

\textbf{Conditional Parameter Generation.} Our approach closely relates to weight-space generation~\cite{han2026survey}, where model parameters are synthesized via hypernetworks or generative models. Early hypernetworks~\cite{ha2016hypernetworks} generate weights from layer-conditioned embeddings and, in some settings, input-dependent signals. However, these methods were primarily developed for Convolutional Neural Networks (CNNs) and Recurrent Neural Networks (RNNs), where the generated weights are comparatively small; the adaptation mechanism is not designed for modern large vision-language models. Subsequent methods scale this idea via checkpoint reconstruction~\cite{peebles2022learning, schurholt2021self, wang2024neural}, but rely on pre-collected adapter banks. Recently, HY-WU~\cite{hywu2026} generates conditional LoRA weights from downstream supervision, but is not designed for temporally grounded video reasoning.

%, avoiding explicit weight supervision. 
%their analyses demonstrate that instance-conditioned generation explores a structured, semantically organized update manifold and that performance gains arise from correct instance–parameter alignment rather than increased capacity per se. 
%Compositional approaches—weight averaging across fine-tuned checkpoints \cite{wortsman2022model} and arithmetic over abstract task vector spaces \cite{ilharco2022editing}—further confirm that the update manifold admits structured, semantically meaningful organization.% amenable to instance-conditioned routing.

% In this paper, we identify the shared-parameter bottleneck as a key limitation in weakly supervised VAD and propose COPRA, which dynamically specializes VLM parameters based on each input.

\begin{figure}
    \centering
    \includegraphics[width=\linewidth]{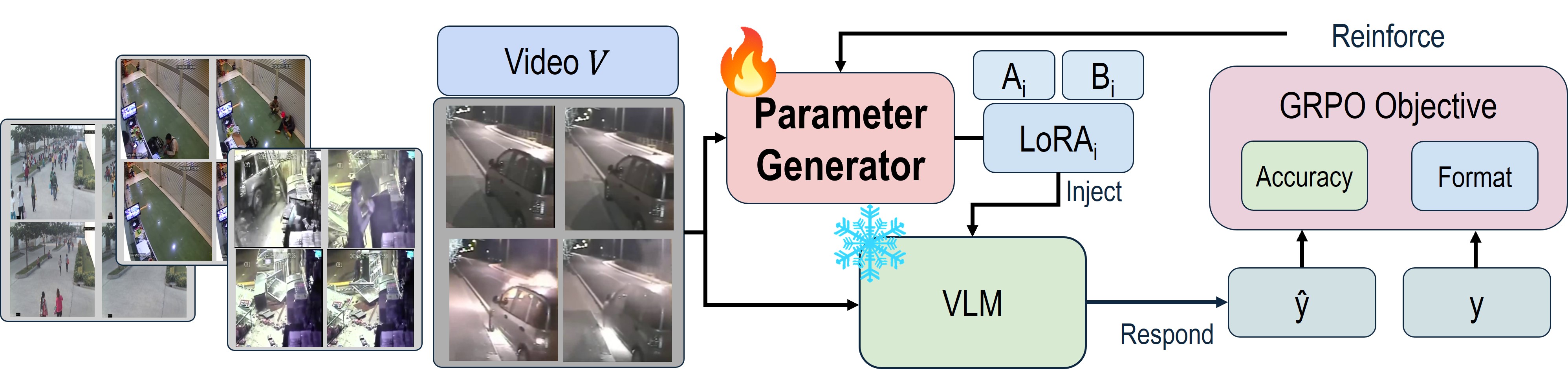}
    \caption{Overview of \textbf{COPRA}, an instance-conditioned parameter generation framework for VAD. The key advantage is a learned functional memory that maps each instance to tailored parameters, enabling improved robustness and cross-domain generalization compared to static adaptation methods.}
    \label{fig:main_figure}
\end{figure}
%nor applied instance-conditioned parameter generation to address it. %Existing VLM-based approaches likewise lack explicit mechanisms for resolving the video-to-frame training--inference mismatch, where models are optimized with coarse video-level supervision yet deployed for dense frame- or segment-level localization. Moreover, robust out-of-distribution generalization across unseen scenes, anomaly types, and datasets remains an open challenge.
%Existing VLM-based VAD methods either avoid weight-based fine-tuning altogether, relying instead on prompts that—whether static or iteratively optimized—remain shared uniformly across all video instances \cite{ye2025vera, zou2025askhint, zanella2024harnessing}, or fine-tune via static shared weight updates, but do so for coarser tasks that relax the frame-level weak supervision constraint \cite{zhang2024holmes, hivau70k, cai2025hiprobe, compactvlm, violencevlm}. 
%We are the first to demonstrate that conditioning VLM parameter updates on input visual context improves frame-level anomaly detection under weak supervision
%with only video-level labels 
%using reinforcement learning. %, and that these gains stem from instance--parameter alignment rather than capacity alone. 
% We explicitly address the video-to-frame training--inference mismatch %, and show this yields additional benefits.

\section{The COPRA framework}

In this section, we present \textbf{COPRA} (\textbf{Co}nditional \textbf{P}a\textbf{R}ameter \textbf{A}daptation), a framework that formulates VAD as an instance-conditioned adaptation problem. To our knowledge, this is the first approach to dynamically specialize VLM behavior for each input video in VAD tasks.

\textbf{Problem Definition}. Given a video $V=\{F_1,\ldots,F_n\}$, the goal is to predict frame-level anomaly labels $y_t \in \{0,1\}$ for each frame $F_t$. In practice, only a video-level label \(y \in \{0,1\}\) is available, where \(y=1\) indicates the presence of at least one anomaly and \(y=0\) denotes a normal video. The objective is to infer fine-grained frame-level scores from this coarse video-level supervision.

\textbf{Overview.} Our COPRA framework adapts VLM parameters on a per-instance basis to overcome the limitations of static, shared adaptations in existing VAD methods, such as supervised fine-tuning (SFT), parameter-efficient tuning (e.g., adapters, LoRA), or prompt optimization. Formally, given a pretrained VLM $f(\cdot; \theta)$, static adaptation learns the global update $\Delta_{\theta_{\text{static}}}$ and produces predictions as
\begin{equation}\label{eq:static}
\hat{y} = f\big(V; \theta + \Delta_{\theta_{\text{static}}}\big).
\end{equation}
As can be seen, \(\Delta_{\theta_{\text{static}}}\) remains fixed across inputs after training---e.g., fixed guiding questions in VERA~\cite{ye2025vera} and fine-tuned LoRA parameters in~\cite{zhang2024holmes}---creating a \textit{shared parameter bottleneck} that leads to a pronounced generalization gap.
Instead, the proposed framework generates instance-conditioned parameter updates:
\begin{equation}\label{eq:dynamic}
\Delta_{\theta}(V) = g_{\varphi_2}\big(\upsilon_{\varphi_1}(V)\big), \quad 
\hat{y} = f\big(V; \theta + \Delta_{\theta}(V)\big).
\end{equation}
Here, \(\upsilon_{\varphi_1}\) and \(g_{\varphi_2}\) denote two sub-networks with parameters \(\varphi_1\) and \(\varphi_2\), respectively. The encoder \(\upsilon_{\varphi_1}\) extracts a compact conditioning signal from the input video \(V\), which is then fed into \(g_{\varphi_2}\) to produce an input-specific parameter update \(\Delta\). This formulation enables adaptive, instance-conditioned modulation of the baseline model and is closely connected to recent advances in weight-space generation~\cite{han2026survey}, where model parameters are dynamically synthesized using hypernetworks or generative models. As illustrated in Figure~\ref{fig:main_problem_visualization}, $\Delta_{\theta}$ serves to adaptively shift the original decision boundary—initially suboptimal for a given instance—toward a more suitable, input-specific configuration, thereby achieving a better VAD performance.

After establishing the foundation, we present the COPRA framework, organized around three key questions: (1) how to design and implement the encoder and parameter generator, (2) how to train them effectively, and (3) how to enable flexible temporal adaptation via a sampling mechanism. The latter is designed to control the granularity at which parameter updates are generated and applied during inference, supporting video-level, segment-level, and hybrid adaptation.

\textbf{Visual Input as Generation Condition.}
To obtain a compact yet representative video feature embedding vectors, we uniformly sample a fixed number of frames (e.g., \(K=8\)). Starting from the first frame, given a video \(V\) with \(n\) frames, we compute a stride \(S = \left\lfloor \frac{n}{K} \right\rfloor\) and construct the subset $\tilde{V} = \{ F_1, F_{1+S}, F_{1+2S}, \dots, F_{1+(K-1)S} \}$. 
The sampled frames $\tilde{V}$ are processed by $\upsilon_{\varphi_1}$, instantiated as a frozen vision encoder (e.g., InternViT in InternVL2-8B~\cite{internvl25}), to produce tokenized visual representations. It allows us to reuse the VLM’s feature extractor, as $\upsilon_{\varphi_1}$ is implemented as a submodule of $f$ in our setting. These representations serve as the sole conditioning signal for  $g_{\varphi_2}$ in our framework, similar to the spirit of~\cite{zhang2024uncovering} and~\cite{hywu2026} for guiding model behavior through targeted adaptation.

\textbf{Conditional Parameter Generation.} Our parameter generator $g_{\varphi_2}$ is a Transformer that maps the concatenated sequence of per-frame visual tokens, i.e., $\upsilon_{\varphi_1}(\tilde{V})$, into tokenized, instance-conditioned LoRA parameters for the language backbone. We initialize the generated LoRA parameters with a learnable global latent prior \(\mathrm{p}_{global}\), which injects dataset-level adaptation patterns into the generated update: $\Delta_{\theta} \leftarrow \Delta_{\theta} + \mathrm{p}_{global}$.
% The parameter generator then refines this prior using video-specific visual evidence to produce the final instance-conditioned LoRA update \(\Delta_{\theta}\).
This is analogous to combining a shared latent component with an input-conditioned residual transformation.
Specifically, the generator $g_{\varphi_2}$ applies factorized self-attention~\cite{hywu2026} over parameter tokens, together with cross-attention to visual features. Intra-layer attention captures dependencies within each backbone block, while inter-layer attention models structural correlations across depth. The resulting representations are then fed into lightweight MLP-based projection heads to generate the LoRA $\mathrm{A}$ and $\mathrm{B}$ matrices. 

We inject these generated LoRA matrices as residual updates into the backbone, covering major projection components such as query–key–value projections, gating modules, and output transformations. We adopt the standard LoRA initialization scheme~\cite{hu2022lora}, where $\mathrm{B}$ is initialized to zero and $\mathrm{A}$ is randomly initialized. This design preserves the pretrained model behavior at the start of training (i.e., a VAD-agnostic initialization) and enables a smooth transition toward instance-conditioned adaptation as learning progresses. The details of the architecture can be found in the appendix. To train the generator, we adopt Group Relative Policy Optimization (GRPO)~\cite{shao2024deepseekmathpushinglimitsmathematical}.

\textbf{End-to-end Reinforcement Learning for COPRA.}
We train COPRA end-to-end by learning a ``policy'' $g_{\varphi_2}$ that maps visual conditioning signals to instance-specific parameter updates. Specifically, let $L$ denote the number of backbone layers, and let $s_l$ denote the number of $r \times d$ parameter tokens required to reconstruct the low-rank adaptation matrices for layer $l$. The generator is defined as $g_{\varphi}: V \rightarrow \Delta_{\theta} \in \mathbb{R}^{L \times s_l \times r \times d}$, where $\Delta_{\theta}$ denotes the generated LoRA parameters.

These parameters are injected into the frozen backbone language model to form $\pi_{\theta + \Delta_{\theta}}$ (e.g., InternLM2 or QwenLM), yielding an instance-adapted policy. Updating only these generated modules while keeping the remaining parameters fixed results in the adapted VLM $f(\cdot; \theta + \Delta_{\theta})$. 
% This formulation is motivated by scenarios where the objective is general—and often non-differentiable—such that explicit supervision over target parameters (e.g., optimal reasoning trajectories or token sequences) is unavailable or prohibitively expensive to obtain. 
The gradients propagate through the generated LoRA weights back to $g_{\varphi_2}$.
%The vision encoder of the backbone remains fixed and provides conditioning features of $\tilde{V}$, while the language model serves as the locus of adaptation through the generated parameters. This design leverages strong pre-trained visual representations while enabling instance-specific specialization via parameter generation.
%Unlike prior parameter generation approaches that rely on supervised regression toward target weights \cite{peebles2022learning, schurholt2021self, wang2024neural}, 
%We cast parameter generation as a reinforcement learning problem. 
%By leveraging RL, the generator directly learns to produce parameters that maximize downstream task performance, implicitly resolving competing optimization signals.
%We train the policy using Group Relative Policy Optimization (GRPO) with rewards defined at the video level. Specifically, we use two rewards:
Details of the reward design, loss functions, and additional implementation choices are provided in the appendix; these components are intentionally kept simple and closely follow the standard GRPO formulation~\cite{shao2024deepseekmathpushinglimitsmathematical}.

\textbf{Flexible Sampling for Temporal Adaptation.} At inference time, we adapt the model to the VAD setting, where the goal is frame-level anomaly scoring. Instead of processing the entire video at once, we partition it into temporally overlapping 10-second segments, each centered at every 16th frame following prior practice~\cite{zanella2024harnessing,ye2025vera}. For each segment, $K=8$ frames are sampled and fed to the parameter generator to produce instance-conditioned LoRA updates, which adapt the VLM backbone for segment-level prediction. The resulting scores from overlapping segments are aggregated to the frame level and further refined using the VERA-style visual retrieval, temporal smoothing, and weighting pipeline~\cite{ye2025vera} to obtain continuous anomaly scores.

Since the sliding-window inference scheme may require frequent re-adaptation, we further investigate flexible temporal adaptation strategies. Specifically, we construct conditioning inputs from broader temporal contexts (e.g., full videos or multi-segment chunks of varying durations) and reuse the resulting parameters to score local segments, enabling flexible temporal adaptation through context-aware parameter sharing. Details are presented in the experimental section (Table~\ref{tab:granularity_ablation}).% In contrast to prior VLM-based VAD pipelines that apply one fixed adaptation to all segments in a video, our method re-parameterizes a frozen backbone according to the temporal context required for prediction.

%Empirically, we observe that adaptation granularity matters. Frequent segment-wise updates improve responsiveness to abrupt temporal changes, while coarser updates yield more stable predictions. This reveals a trade-off between temporal sensitivity and adaptation stability, suggesting update frequency as an important design axis for instance-conditioned VAD systems which we explore more in Table \ref{tab:granularity_ablation}.
%For instance, VERA \cite{ye2025vera} uses the fixed prompt across all segments in a video. Instead, our model performs \emph{test-time conditional adaptation}: the language model is re-parameterized for each temporal segment based on its local visual content. This allows the model to respond differently to heterogeneous contexts (e.g., parking lots, hallways, roads) and to distinct anomaly patterns occurring within the same video.

%After obtaining segment-level scores, we adopt the post-processing procedure of \cite{ye2025vera}. For each segment, we retrieve visually similar segments from the same video and use their anomaly scores to refine the current prediction, thereby incorporating broader scene context. We then apply temporal Gaussian smoothing to convert segment predictions into final frame-wise anomaly scores. %Keeping this post-processing fixed isolates the contribution of our method to local segment-level anomaly understanding rather than auxiliary heuristics.

\section{Experiments}

% We evaluate whether COPRA improves VAD performance, robustness, and generalization compared to standard parameter-efficient adaptation methods. Our experiments aim to answer three questions:  First, we ask whether dynamic, instance-conditioned adaptation provides a stronger paradigm than static adaptation methods for VAD; Second, we examine whether this advantage holds in a parameter-efficient setting by comparing neural generation against static LoRA-based GRPO baselines under matched training and adaptation conditions. Third, we probe the generalization boundary of instance-conditioned parameter generation by evaluating whether it transfers to unseen tasks beyond VAD and shifted data distributions.
We evaluate COPRA in terms of performance, robustness, and generalization. Specifically, we address three questions: (i) whether instance-conditioned adaptation outperforms static adaptation approaches; (ii) whether this advantage persists under parameter-efficient settings compared to LoRA-based GRPO baselines; and (iii) whether the learned adaptation generalizes to unseen tasks and distribution shifts both within and beyond VAD.
% \end{itemize}

\subsection{Experimental Settings}

\textbf{In-Domain Datasets.} We train our method on two standard VAD benchmarks: UCF-Crime \cite{sultani2018real} and XD-Violence \cite{Wu2020not}. Following common weakly supervised VAD practice, we cast evaluation as frame-level binary anomaly classification (normal vs.\ abnormal).
%UCF-Crime consists of real-world surveillance footage with 1,610 training videos and 290 testing videos. The dataset contains 14 anomaly categories (e.g., Abuse, Assault, Burglary, Road Accidents, and Vandalism) together with normal videos. Following common weakly supervised VAD practice, we cast evaluation as binary anomaly detection (normal vs.\ abnormal), ignoring fine-grained category labels. XD-Violence contains 4,754 videos, with 800 reserved for testing, collected from diverse sources including CCTV footage, movies, and online platforms such as YouTube. Although XD-Violence defines 7 anomaly categories, we again evaluate under the binary anomaly detection setting.

\begin{table*}[t]
\centering
\tablestyle{8pt}{1.2}

\caption{
In-domain comparison with prior VAD methods on UCF-Crime and XD-Violence.
Among explainable approaches, COPRA consistently improves over its frozen baseline and outperforms strong baselines such as VERA~\cite{ye2025vera}. On UCF-Crime, it raises InternVL2-8B from 84.31 to 87.14 AUC, and on XD-Violence from 74.32 to 76.52 AP.
}
\label{tab:main_results_combined}

\begin{minipage}[t]{0.49\textwidth}
\centering
\small

\textbf{(a) UCF-Crime}

\begin{tabular}{lc}
\shline
Method & AUC (\%) \\
\shline

\multicolumn{2}{c}{\textit{Prior Unexplainable VAD Methods}} \\
\shline
RTFM~\cite{tian2021weakly}       & 84.30 \\
SSRL~\cite{ssrl}                 & 87.43 \\
MSL~\cite{li2022self}            & 85.62 \\
CLIP-TSA~\cite{joo2023clip}      & 87.58 \\
MGFN~\cite{chen2023mgfn}         & 86.98 \\
OVVAD~\cite{wu2024open}          & 86.40 \\

\shline
\multicolumn{2}{c}{\textit{Prior Explainable VAD Methods}} \\
\shline
LAVAD~\cite{zanella2024harnessing}        & 80.28 \\
Holmes-VAD~\cite{zhang2024holmes}         & 84.61 \\
VADor~\cite{Lv2024VideoAD}                & 85.90 \\
LLAVA-1.5~\cite{liu2024improved}          & 72.84 \\
VERA (InternVL2-8B)~\cite{ye2025vera}     & 86.55 \\

\shline
\multicolumn{2}{c}{\textit{Frozen Baseline}} \\
\shline
\rowcolor{baselinecolor}
Qwen2-VL-7B-Instruct (Frozen)              & 81.31 \\
\rowcolor{baselinecolor}
InternVL2-8B (Frozen)                      & 84.31 \\

\shline
\multicolumn{2}{c}{\textit{COPRA}} \\
\shline
\rowcolor{nvidiagreen}
Qwen2-VL-7B-Instruct                  & \textbf{86.75} \\
\rowcolor{nvidiagreen}
InternVL2-8B                          & \textbf{87.14} \\
%\rowcolor{nvidiagreen}
%InternVL3-8B                          & \textbf{87.33} \\
\shline
\end{tabular}
\end{minipage}
\begin{minipage}[t]{0.49\textwidth}
\centering
\small
\textbf{(b) XD-Violence}

\begin{tabular}{lc}
\shline
Method & AP (\%) \\
\shline

\multicolumn{2}{c}{\textit{Prior Unexplainable VAD Methods}} \\
\shline
%Wu et al.~\cite{Wu2020not}              & 78.64 \\
RTFM~\cite{tian2021weakly}              & 77.81 \\
S3R~\cite{10.1007/978-3-031-19778-9_42} & 80.26 \\
MSL~\cite{li2022self}                   & 78.58 \\
CLIP-TSA~\cite{joo2023clip}             & 82.19 \\
MGFN~\cite{chen2023mgfn}                & 80.11 \\
OVVAD~\cite{wu2024open}                 & 66.53 \\
\shline
\multicolumn{2}{c}{\textit{Prior Explainable VAD Methods}} \\
\shline
LAVAD ~\cite{zanella2024harnessing}         & 62.01 \\
Holmes-VAD~\cite{zhang2024holmes}           & 84.96 \\
%ZS CLIP~\cite{zanella2024harnessing}        & 17.83 \\
%ZS ImageBind-I~\cite{zanella2024harnessing} & 27.25 \\
ZS ImageBind-V~\cite{zanella2024harnessing} & 25.36 \\
LLAVA-1.5~\cite{liu2024improved}            & 50.26 \\
VERA (InternVL2-8B)~\cite{ye2025vera}       & 70.54 \\
\shline
\multicolumn{2}{c}{\textit{Frozen Baseline}} \\
\shline
\rowcolor{baselinecolor}
Qwen2-VL-7B-Instruct (Frozen)                       & 64.36 \\
\rowcolor{baselinecolor}
InternVL2-8B (Frozen)                               & 74.32 \\

\shline
\multicolumn{2}{c}{\textit{COPRA}} \\
\shline
\rowcolor{nvidiagreen}
Qwen2-VL-7B-Instruct                           & \textbf{73.96} \\
\rowcolor{nvidiagreen}
InternVL2-8B                                  & \textbf{76.52} \\
\shline
\end{tabular}
\end{minipage}
\end{table*}

\begin{table}[t]
\centering
\tablestyle{12pt}{1.2}

\caption{
COPRA consistently outperforms frozen VLM baselines across models and in-domain datasets. For fair comparison, all models use VERA’s three-stage inference—(Step 1) initial scoring, (Step 2) retrieval refinement, and (Step 3) temporal smoothing/weighting—and we compare identical backbones with and without COPRA. COPRA improves AUC/AP at every stage for Qwen2-VL-7B-Instruct and InternVL2-8B, demonstrating gains beyond post-processing alone.
}
\label{tab:pg_all_models_all_datasets}

% \resizebox{0.86\textwidth}{!}{%
\begin{tabular}{ll>
{\columncolor{baselinecolor}}c>{\columncolor{nvidiagreen}}c>{\columncolor{baselinecolor}}c>{\columncolor{nvidiagreen}}c}

\multirow{2}{*}{Model} 
& \multirow{2}{*}{Stage}
& \multicolumn{2}{c}{\textbf{UCF-Crime}}
& \multicolumn{2}{c}{\textbf{XD-Violence}} \\
& 
& Frozen & \textbf{+ COPRA}
& Frozen & \textbf{+ COPRA} \\
\shline

\multirow{3}{*}{Qwen2-VL-7B-Instruct}
& Step 1 & 59.36 & \textbf{69.73} & 54.01 & \textbf{63.09} \\
& Step 2 & 77.91 & \textbf{84.69} & 62.01 & \textbf{72.60} \\
& Step 3 & 81.31 & \textbf{86.75} & 64.36 & \textbf{73.96} \\

\shline

\multirow{3}{*}{InternVL2-8B}
& Step 1 & 64.95 & \textbf{71.49} & 61.53 & \textbf{67.24} \\
& Step 2 & 81.74 & \textbf{85.21} & 72.83 & \textbf{75.80} \\
& Step 3 & 84.31 & \textbf{87.14} & 74.32 & \textbf{76.52} \\

\end{tabular}%
% }

\end{table}

\begin{table}[t]
\centering
\tablestyle{8pt}{1.2}

\caption{
Ablation on UCF-Crime comparing COPRA with static adaptation under GRPO training. The frozen baseline uses the standard VLM pipeline; static adaptation applies shared LoRA with GRPO; COPRA adds instance-conditioned LoRA to the frozen baseline under the same GRPO setting. COPRA consistently outperforms both baselines across all stages, indicating that the gains come from adaptive conditioning rather than GRPO alone.
}
\label{tab:pg_ablation}

\resizebox{\textwidth}{!}{%
\begin{tabular}{p{6.4cm}>{\columncolor{baselinecolor}}cc>{\columncolor{nvidiagreen}}c}
\multirow{2}{*}{Inference Stage}
& Frozen & GRPO & \textbf{Frozen Baseline + COPRA}\\
& Baseline & Static Adaptation & \textbf{Dynamic Adaptation}\\
\shline

Initial (Step 1)
& 59.36 & 61.34 & \textbf{69.73}\\

Initial + Retrieval (Step 2)
& 77.91 & 79.51 & \textbf{84.69}\\

Initial + Retrieval + Smoothing + Weighting (Step 3)
& 81.31 & 83.00 & \textbf{86.75}\\

\end{tabular}
}

\end{table}

\begin{figure}
    \centering
    \includegraphics[width=.94\textwidth]{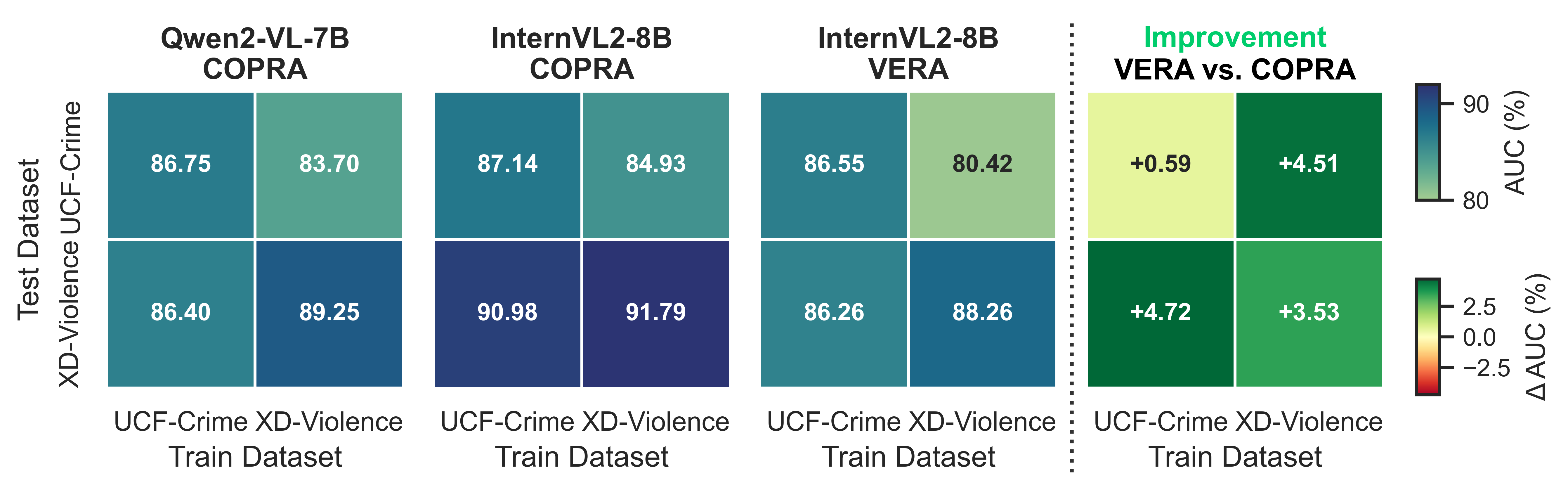}
    \caption{AUC (\%) performance for cross-dataset generalization across train--test dataset pairs.
    From left to right: COPRA with Qwen2-VL-7B-Instruct, COPRA with InternVL2-8B, VERA with InternVL2-8B, and the AUC improvement of COPRA over VERA~\cite{ye2025vera}.}
    \label{fig:cross_dataset_generalization}
\end{figure}

\begin{table}[!htbp]
\centering
\tablestyle{8pt}{1.2}

\caption{
Cross-domain dense captioning results on VRU-Accident~\cite{Kim_2025_ICCV}. Prec., Rec., and F1 denote ROUGE-1/2/L scores. COPRA improves semantic metrics (SPICE, METEOR, COMET) and boosts ROUGE recall and F1 for higher-order structures (bigrams and LCS), indicating better recovery of accident-relevant content and stronger alignment with ground-truth descriptions.
}
\label{tab:vru_accident_generalization_dense_captioning}

\resizebox{\textwidth}{!}{%
\large
\begin{tabular}{l|cccccccccccc}
\multirow{2}{*}{Model \textbackslash \ Metric}
& \multicolumn{3}{c}{Semantic Metrics}
& \multicolumn{3}{c}{ROUGE-1}
& \multicolumn{3}{c}{ROUGE-2}
& \multicolumn{3}{c}{ROUGE-L} \\
& SPICE & METEOR & COMET
& Prec. & Rec. & F1
& Prec. & Rec. & F1
& Prec. & Rec. & F1 \\
\shline

\rowcolor{baselinecolor}
InternVL2-8B 
& 0.139 & 0.249 & 0.694
& \textbf{0.398} & 0.431 & \textbf{0.410}
& \textbf{0.091} & 0.098 & 0.093
& \textbf{0.201} & 0.220 & 0.208 \\

\rowcolor{nvidiagreen}
\textbf{+ COPRA}
& \textbf{0.149} & \textbf{0.273} & \textbf{0.701}
& 0.359 & \textbf{0.489} & \textbf{0.410}
& 0.090 & \textbf{0.123} & \textbf{0.103}
& 0.186 & \textbf{0.254} & \textbf{0.212} \\

\shline
Improvement ($\uparrow$)
& {\color{mygreen}{+0.010}} & {\color{mygreen}{+0.024}} & {\color{mygreen}{+0.007}}
& {\color{darkred}{-0.039}} & {\color{mygreen}{+0.058}} & {\color{mygreen}{+0.000}}
& {\color{darkred}{-0.001}} & {\color{mygreen}{+0.025}} & {\color{mygreen}{+0.010}}
& {\color{darkred}{-0.015}} & {\color{mygreen}{+0.034}} & {\color{mygreen}{+0.004}} \\

\end{tabular}%
}

\end{table}

\begin{table}[t]
\centering
\caption{
Cross-domain multiple-choice video question answering results on VRU-Accident~\cite{Kim_2025_ICCV}. Compared with the frozen InternVL2-8B baseline, COPRA improves overall accuracy by 1.35 points, with the largest gain of 6.67 points on DoTA.
}
\label{tab:vru_accident_generalization}
\tablestyle{8pt}{1.2}
% \resizebox{\linewidth}{!}{%
\begin{tabular}{l|cccc|c}
\multirow{2}{*}{Model \textbackslash \ Dataset}
& DADA & CAP & DoTA & MANU & \textbf{Overall} \\
& 1,338 & 1,722 & 600 & 2,340 & 6,000 \\
\shline

\rowcolor{baselinecolor}
InternVL2-8B
& 48.58 & \textbf{52.50} & 44.83 & 47.86 & 49.05 \\

\rowcolor{nvidiagreen}
\textbf{InternVL2-8B + COPRA}
& \textbf{49.70} & 50.41 & \textbf{51.50} & \textbf{50.51} & \textbf{50.40} \\

\shline
improvement ($\uparrow$)
& 1.12 & -2.09 & 6.67 & 2.65 & 1.35 \\
\end{tabular}%
% }

\end{table}

\begin{figure}[t]
\centering
\tablestyle{8pt}{1.2}
\includegraphics[width=\linewidth]{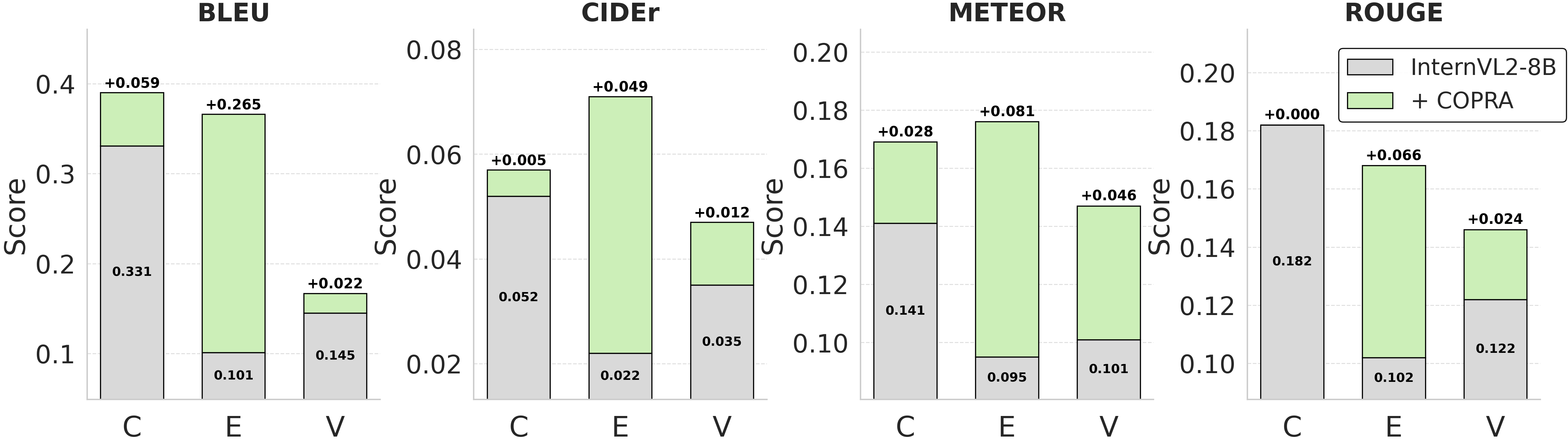}
\caption{Cross-domain evaluation of natural language explanation quality on HIVAU-70K~\cite{hivau70k}. C, E, and V denote clip-, event-, and video-level evaluation, respectively. Compared with the InternVL2-8B baseline, COPRA improves most semantic and lexical overlap metrics across all temporal levels, including BLEU, CIDEr, METEOR, and ROUGE, demonstrating strong cross-domain generalization for natural language explanation generation.}
\label{fig:generalization_hivau70k_transposed}

% \resizebox{\columnwidth}{!}{%
\end{figure}

\begin{table}[!htbp]
\centering
\tablestyle{8pt}{1.2}

\caption{
Effect of temporal granularity on UCF-Crime Test AUC for Qwen2-VL-7B.
%COPRA is evaluated by generating parameters at different temporal scales, from 10s segments to full-video adaptation.
The 120s chunk setting achieves the best performance, improving over video-level generation by 0.65 AUC.
%Best results are bolded.
}
\label{tab:granularity_ablation}

% \resizebox{\linewidth}{!}{%
\begin{tabular}{l|ccc>{\columncolor{nvidiagreen}}ccc}
\multirow{2}{*}{Metric \textbackslash \ Granularity}
& 10s & 30s & 60s & 120s & 240s & Video \\
& Segment & Chunk & Chunk & \textbf{Chunk} & Chunk & Full \\
\shline

AUC (\%)
& 86.51 & 86.63 & 86.48 & \textbf{87.40} & 86.56 & 86.75 \\

Improvement over video-level generation
& -0.24 & -0.12 & -0.27 & \textbf{+0.65} & -0.19 & -- \\

\end{tabular}%
% }

\end{table}

\textbf{Cross-Domain Datasets.} To evaluate whether COPRA generalizes beyond surveillance VAD, we conduct zero-shot evaluation, without fine-tuning, on two out-of-domain benchmarks with distinct objectives. 
VRU-Accident~\cite{Kim_2025_ICCV}, built from DADA~\cite{dada2000}, DoTA~\cite{9712446}, and additional curated CAP and MANU videos, focuses on traffic safety scenarios involving vulnerable road users and supports both multiple-choice video question answering and dense captioning. We further evaluate on HIVAU-70K~\cite{hivau70k}, an open-ended anomaly-centric video understanding benchmark. 
Notably, COPRA is trained only on UCF-Crime and directly evaluated on these datasets without additional fine-tuning, enabling a rigorous cross-domain generalization assessment. 
% To test whether the learned parameter generator transfers beyond surveillance anomaly detection, without any finetuning, we additionally evaluate on two out-of-domain benchmarks with fundamentally different objectives. VRU-Accident \cite{Kim_2025_ICCV} is a traffic safety benchmark designed to evaluate video understanding in accident scenarios involving vulnerable road users such as pedestrians and cyclists. The benchmark is constructed from four sources: DADA \cite{dada2000}, DoTA \cite{9712446}, and additional manually captioned videos curated by the authors.
% We evaluate zero-shot transfer on (1) multiple-choice video question answering,
%  and (2) dense captioning, where the model generates natural-language accident descriptions. 
% We evaluate general video understanding on HIVAU-70K \cite{hivau70k}, an open-ended, anomaly-focused question answering benchmark built from clips, events, and videos. Note that COPRA is trained only on UCF-Crime and evaluated directly on these cross-domain datasets to ensure a rigorous generalization assessment. 

\textbf{Metrics.} We follow standard weakly supervised VAD protocols, reporting frame-level AUC (Area Under the Curve, \%) on UCF-Crime and Average Precision (AP, \%) on XD-Violence, consistent with common practice in the VAD literature.
% as it better captures precision–recall behavior under class imbalance, where abnormal frames are relatively over-represented compared to UCF-Crime.
For VRU-Accident multiple-choice video QA, we report classification accuracy (\%).
For dense captioning on VRU-Accident, we use semantic metrics (SPICE~\cite{Anderson2016SPICESP}, METEOR~\cite{lavie-agarwal-2007-meteor}, COMET~\cite{rei-etal-2020-comet}) and ROUGE~\cite{lin-2004-rouge} for lexical overlap.
For HIVAU-70K, we report BLEU~\cite{papineni-etal-2002-bleu}, CIDEr~\cite{Vedantam2014CIDErCI}, METEOR, and ROUGE at the clip-, event-, and video-level.

\subsection{Comparison to State-of-the-Art Methods}

%\begin{table*}[!htbp]
%\centering
%\resizebox{0.8\textwidth}{!}{%
%\begin{minipage}[t]{0.48\textwidth}
%\centering
%\small
%\textbf{(a) UCF-Crime}
%
%\vspace{0.3em}

%\begin{tabular}{llcc}
%\toprule
%Model & Stage & Baseline & + NG \\
%\midrule
%\multirow{3}{*}{Qwen2-VL-7B-Instruct}
%& Step 1 & 59.36 & \textbf{69.73} \\
%& Step 2 & 77.91 & \textbf{84.69} \\
%& Step 3 & 81.31 & \textbf{86.75} \\
%\midrule
%\multirow{3}{*}{InternVL2-8B}
%& Step 1 & 64.95 & \textbf{71.49} \\
%& Step 2 & 81.74 & \textbf{85.21} \\
%& Step 3 & 84.31 & \textbf{87.14} \\
%\bottomrule%
%\end{tabular}
%\end{minipage}
%\hfill
%\begin{minipage}[t]{0.48\textwidth}
%\centering
%\small
%\textbf{(b) XD-Violence}

%\vspace{0.3em}

%\begin{tabular}{llcc}
%\toprule
%Model & Stage & Baseline & + NG \\
%\midrule
%\multirow{3}{*}{Qwen2-VL-7B-Instruct}
%& Step 1 & 58.82 & \textbf{72.85} \\
%& Step 2 & 76.91 & \textbf{88.22} \\
%& Step 3 & 77.97 & \textbf{89.25} \\
%\midrule
%\multirow{3}{*}{InternVL2-8B}
%& Step 1 & 71.78 & \textbf{77.56} \\
%& Step 2 & 89.53 & \textbf{91.06} \\
%& Step 3 & 90.63 & \textbf{91.79} \\
%\bottomrule
%\end{tabular}
%\end{minipage}

%\vspace{0.5em}
%\caption{Performance comparison of frozen baselines and instance-conditioned neural generation (NG) across inference stages on UCF-Crime and XD-Violence.}
%\label{tab:pg_all_models_all_datasets}
%\end{table*}

Table~\ref{tab:main_results_combined} shows that COPRA achieves improved performance over prior explainable methods, consistently across architectures. Notably, it approaches strong non-explainable baselines while employing a fundamentally different instance-conditioned adaptation mechanism and providing interpretable reasoning. Holmes-VAD is included among prior explainable VAD methods due to its strong performance. However, because it relies on fine-grained temporal annotations, it is not directly comparable under the weakly supervised VAD setting.
%Rather than relying on static parameter updates or task-specific fine-tuning, neural generation enables instance-conditioned adaptation, which we show in subsequent sections leads to strong cross-dataset generalization.
Table~\ref{tab:pg_all_models_all_datasets} further shows that these gains generalize across backbones (Qwen2-VL-7B-Instruct vs InternVL2-8B), datasets (UCF-Crime and XD-Violence), with consistent improvements at each inference step when COPRA is leveraged.
\newcommand{\improve}[1]{$_{\color{black}\uparrow #1}$}

\subsection{Ablation Studies}

\textbf{Instance-Specific Parameters Outperform Static Adaptation}. To further validate COPRA, we evaluate three configurations: (1) a \textit{Frozen Baseline} with no adaptation; (2) \textit{Static Adaptation with GRPO}, which applies shared, dataset-specific LoRA parameters (Eq.~\ref{eq:static}) trained with GRPO; and (3) \textit{Frozen Baseline with COPRA}, which uses instance-conditioned LoRA parameters (Eq.~\ref{eq:dynamic}) trained with GRPO. As shown in Table~\ref{tab:pg_ablation}, \textit{Frozen Baseline with COPRA} delivers substantial gains, improving AUC by 5.4 points over the frozen baseline and by 3.75 points over static adaptation with GRPO. While static adaptation applies a shared update across all inputs, leading to a compromised solution, COPRA performs on-the-fly, input-conditioned adaptation that better handles conflicting patterns such as anomaly types and scene contexts.

\textbf{Instance-Specific Parameters Generalize In-Domain Datasets}. A key question is whether improved in-domain performance reflects over-specialization. To test this, we train COPRA on UCF-Crime and evaluate on both UCF-Crime and XD-Violence, and conversely train on XD-Violence and evaluate on both datasets. As shown in Figure~\ref{fig:cross_dataset_generalization}, COPRA consistently improves over baselines and outperforms VERA across all train–test combinations, indicating strong cross-dataset generalization.

\textbf{Instance-Specific Parameters Generalize Cross-Domain Datasets}. We further evaluate cross-domain generalization by directly applying the COPRA model trained on UCF-Crime without additional fine-tuning. On VRU-Accident, COPRA improves both dense captioning quality (Table~\ref{tab:vru_accident_generalization_dense_captioning}) and zero-shot multiple-choice VideoQA accuracy (Table~\ref{tab:vru_accident_generalization}), suggesting strong transfer to safety-critical video understanding under substantial domain shift. Notably, the improvements indicate that instance-conditioned adaptation captures transferable abnormality cues beyond dataset-specific patterns learned during training. On HIVAU-70K, COPRA also improves natural-language explanations (Figure~\ref{fig:generalization_hivau70k_transposed}), yielding consistent gains across BLEU, CIDEr, METEOR, and ROUGE. These results further demonstrate that COPRA generalizes beyond anomaly scoring to broader cross-domain video understanding and reasoning tasks.
%Notably, these comparisons are controlled for parameter count, indicating that the gains arise from how parameters are adapted rather than how many are used. 
%A fixed LoRA adapter must encode a single set of parameters shared across all inputs. However, video anomaly detection involves heterogeneous and potentially conflicting objectives across videos (e.g., varying anomaly types, motion dynamics, and scene contexts). 
% We hypothesize that such fixed adapters are susceptible to cross-instance interference, which limits representational capacity.

\textbf{Flexible Sampling for Temporal Adaptation in COPRA.}
% COPRA decouples the temporal scale for parameter generation from the scale for anomaly scoring.
Existing VAD approaches typically use heavily overlapping sliding windows (e.g., 10 seconds) with the same static parameters applied to all windows. In contrast, COPRA adapts to different temporal stages by generating chunk-specific parameters. Specifically, we divide videos into non-overlapping 30-, 60-, or 120-second chunks, sample \(K=8\) frames per chunk, and generate low-rank parameters once per chunk. Each 10-second segment is then scored using the parameters of its corresponding chunk, amortizing parameter generation while preserving broader temporal context. As shown in Table~\ref{tab:granularity_ablation}, COPRA remains robust across temporal granularities, with the best achieved from the 120-second chunks.

%\begin{table}
%\centering
%\small
%\setlength{\tabcolsep}{5pt}
%\renewcommand{\arraystretch}{1.1}
%\begin{tabular}{lcc}
%\toprule
%Neural Generation Granularity & Test AUC (\%) & $\Delta$ \\
%\midrule
%%Video-Level                  & 86.75 & -- \\
%Segment-Level                & 86.51 & -0.24 \\
%Chunk-Level (30s)            & 86.63 & -0.12 \\
%Chunk-Level (60s)            & 86.48 & -0.27 \\
%Chunk-Level (120s)           & \textbf{87.40} & +0.65 \\
%Chunk-Level (240s)           & 86.56 & -0.19 \\
%\bottomrule
%\end{tabular}
%\vspace{0.25em}
%\caption{Effect of neural generation granularity on UCF-Crime test AUC for Qwen2-VL-7B-Instruct after one epoch of training. Deltas are reported relative to video-level neural generation. Moderate temporal chunking (120s) achieves the best result.}
%\label{tab:granularity_ablation}
%\end{table}

\begin{table}
\centering
\tablestyle{8pt}{1.2}
\caption{
Effect of the learnable global latent in InternVL2-8B + COPRA on UCF-Crime after one epoch of training.
%The frozen baseline uses the same VLM inference pipeline without generated parameters, while COPRA variants generate instance-conditioned adapters either with or without learnable global latent priors.
Both COPRA variants improve over the frozen baseline across all inference stages, while including the learnable global latent during training achieves the strongest performance.
}
\label{tab:ucf_global_priors}

% \resizebox{\linewidth}{!}{%
\begin{tabular}{l>{\columncolor{baselinecolor}}cc>{\columncolor{nvidiagreen}}c}
\multirow{2}{*}{Inference Stage}
& Frozen & COPRA & \textbf{COPRA} \\
& Baseline & No Global Latent & Learnable Global Latent \\
\shline

Initial (Step 1)
& 64.95 & \textbf{71.79} & 71.49 \\

Initial + Retrieval (Step 2)
& 81.74 & 84.43 & \textbf{85.21} \\

Initial + Retrieval + Smoothing + Weighting (Step 3)
& 84.31 & 86.39 & \textbf{87.14} \\

\end{tabular}%
% }

\end{table}

\textbf{Role of the Learnable Global Latent.}
Table~\ref{tab:ucf_global_priors} shows that COPRA consistently outperforms the frozen baseline both with and without the learnable global latent $\mathrm{p}_{global}$ during training. Removing \(\mathrm{p}_{global}\) eliminates the dataset-level prior, forcing the generator to rely solely on video-conditioned features. 
%As can be shown, a learned \(\mathrm{p}_{global}\) provides structural priors for the generated LoRA updates, initializing the generator in a more informative region of the parameter space before refinement with video-specific evidence. 
Including $\mathrm{p}_{global}$ improves post-processed results by approximately \(0.75\%\) AUC.

\subsection{Qualitative Results and Case Studies}

\begin{figure}
    \centering

    \begin{subfigure}[t]{0.47\textwidth}
        \centering
        \includegraphics[width=\linewidth,height=0.85\linewidth,keepaspectratio]{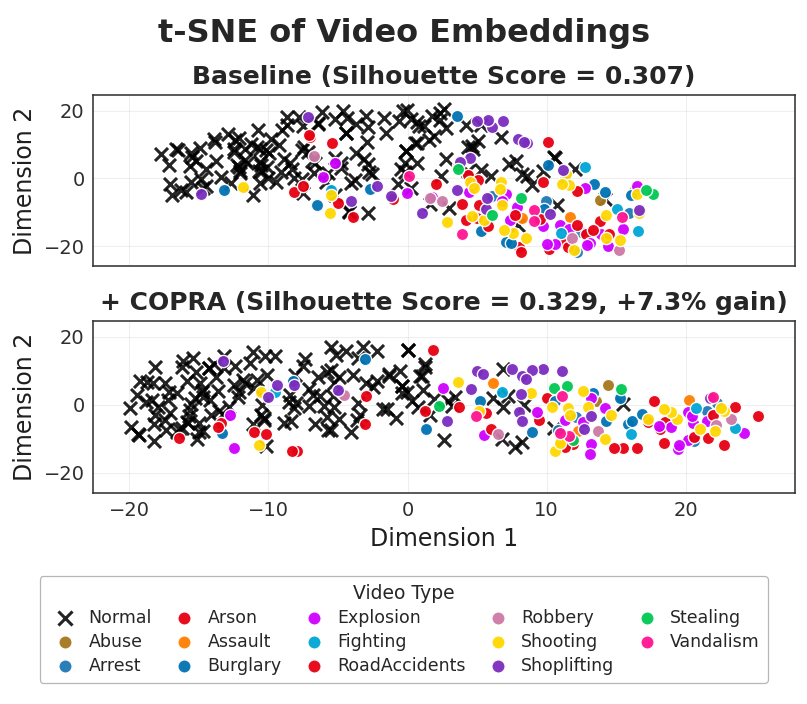}
        \label{fig:tsne_embeddings}
    \end{subfigure}
    \hfill
    \begin{subfigure}[t]{0.52\textwidth}
        \centering        
        \includegraphics[width=\linewidth,height=0.85\linewidth,keepaspectratio]{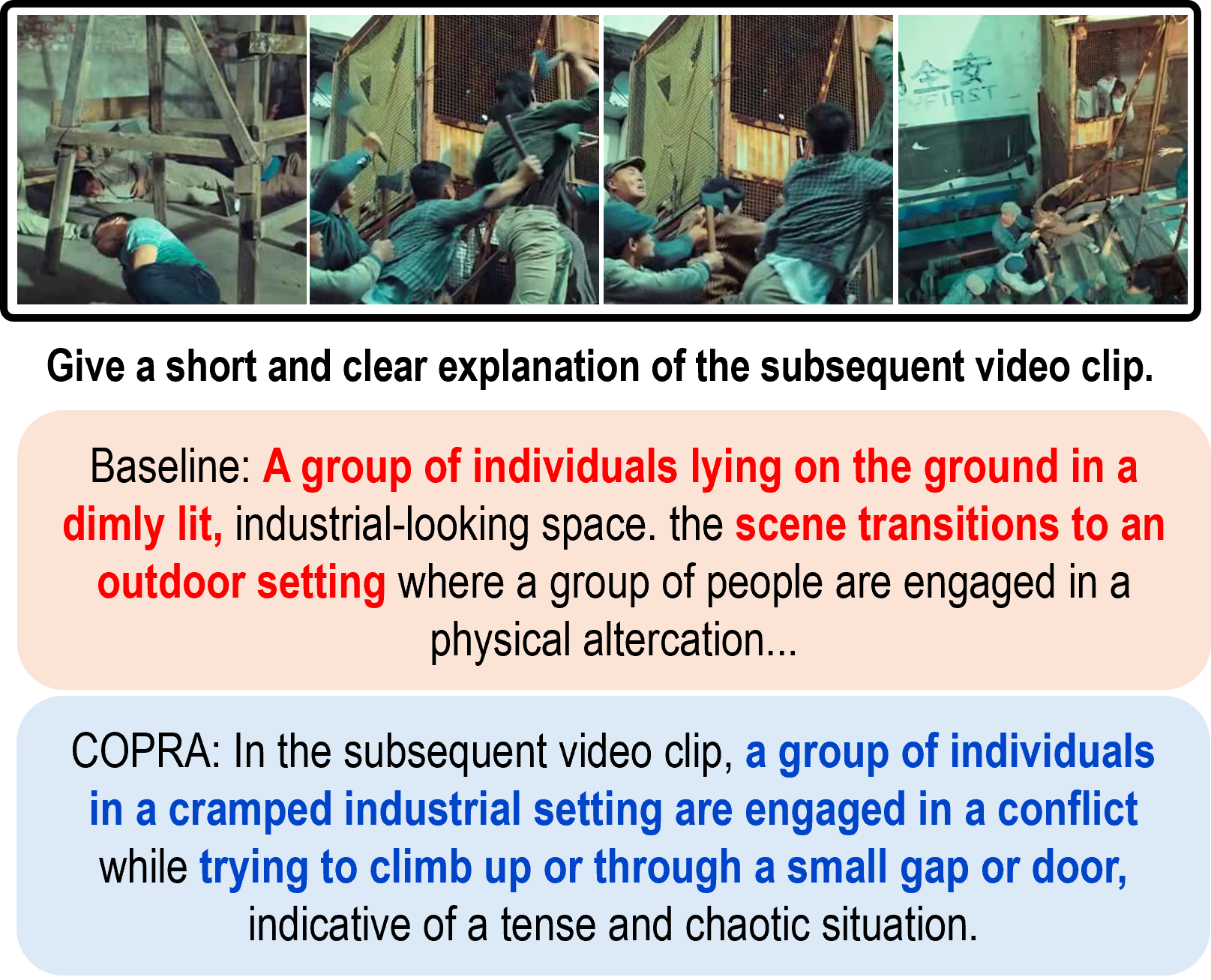}
        \label{fig:qualitative_1}
    \end{subfigure}

    \caption{
Qualitative analysis of InternVL2-8B + COPRA representations and responses. 
Left: Under shared t-SNE axes, COPRA produces more discriminative embeddings, with normal and anomalous videos more clearly separated; we quantify this separation using the silhouette score, where higher values indicate better clustering quality.
Right: COPRA generates anomaly-relevant descriptions with behaviorally salient cues, whereas the baseline mainly produces generic scene descriptions.
    }
    \label{fig:qualitative_analysis}
\end{figure}

Figure~\ref{fig:qualitative_analysis} compares the representation space (left) and generated descriptions of the baseline model and COPRA (right). COPRA separates normal and anomalous videos more clearly, including categories such as shoplifting that remain close to normal videos in the baseline space. We quantify class separation using the silhouette score~\cite{ROUSSEEUW198753}, computed on the two-dimensional t-SNE embeddings.
% For each video $V_i$, the silhouette coefficient is $s(V_i)=\frac{b(V_i)-a(V_i)}{\max\{a(V_i),b(V_i)\}}$, where $a(V_i)$ denotes the mean intra-class distance and $b(V_i)$ the mean distance to the nearest competing class. The final score averages $s(V_i)$ over all videos. 
%Higher silhouette scores indicate better class separation. COPRA increases the silhouette score from 0.307 to 0.329, a 7.3\% relative gain, suggesting that COPRA produces a representation space in which anomalous videos are more distinguishable from normal videos.
This difference is reflected in the generated descriptions: the baseline produces a fragmented narrative that abruptly shifts between unrelated scenes, whereas COPRA generates a more coherent anomaly-focused description by consistently capturing the industrial setting, ongoing conflict, and tense attempts to escape through a narrow opening. We provide details in the appendix.

\section{Conclusion}
We identify several fundamental limitations of existing VAD approaches, including the \textit{shared parameter bottleneck}, where static shared adaptations force heterogeneous requirements into a single compromised solution, as well as the mismatch between training and inference in both data distribution and model configuration. We introduced COPRA, an instance-conditioned parameter generation framework that adapted a frozen VLM to each video by generating low-rank weights trained with reinforcement learning under weak video-level supervision. COPRA improved robustness to diverse scenes, anomaly types, and temporal dynamics. Across both in-domain and cross-domain evaluations, COPRA consistently outperformed frozen and static adaptation baselines and could potentially serve as an effective weight space generation
mechanism for foundation models, enabling scalable, adaptive, and context-aware
video understanding.
% NEEDED FOR NEURIPS CHECKLIST
Limitations include limited interpretability of generated LoRA weights, lack of theoretical grounding, and additional computational overhead introduced by repeated response sampling during GRPO training, all of which warrant future investigation.

% These results suggest that reinforcement learning provides an effective signal for training parameter generators when target weights are unavailable. 
% Limitations include limited interpretability of generated weights, lack of theoretical grounding, and the computational cost of repeated response sampling during GRPO training.

\bibliographystyle{plainnat}
\bibliography{references}

@inproceedings{zou2026unlocking,
  title={Unlocking vision-language models for video anomaly detection via fine-grained prompting},
  author={Zou, Shu and Tian, Xinyu and Wesemann, Lukas and Waschkowski, Fabian and Yang, Zhaoyuan and Zhang, Jing},
  booktitle={WACV},
  pages={4223--4233},
  year={2026}
}

@inproceedings{sultani2018real,
  title={Real-world anomaly detection in surveillance videos},
  author={Sultani, Waqas and Chen, Chen and Shah, Mubarak},
  booktitle={CVPR},
  pages={6479--6488},
  year={2018}
}

@inproceedings{
    zhang2024uncovering,
    title={Uncovering Latent Chain of Thought Vectors in Large Language Models},
    author={Jason Zhang and Scott W Viteri},
    booktitle={Workshop on Neural Network Weights as a New Data Modality},
    year={2025},
    url={https://openreview.net/forum?id=ICuIdJzBPm}
}

@misc{han2026survey,
      title={A Survey of Weight Space Learning: Understanding, Representation, and Generation}, 
      author={Xiaolong Han and Zehong Wang and Bo Zhao and Binchi Zhang and Jundong Li and Damian Borth and Rose Yu and Haggai Maron and Yanfang Ye and Lu Yin and Ferrante Neri},
      year={2026},
      eprint={2603.10090},
      archivePrefix={arXiv},
      primaryClass={cs.LG},
      url={https://arxiv.org/abs/2603.10090}, 
}

@InProceedings{huang2025ex,
  title = 	 {Ex-{VAD}: Explainable Fine-grained Video Anomaly Detection Based on Visual-Language Models},
  author =       {Huang, Chao and Shi, Yushu and Wen, Jie and Wang, Wei and Xu, Yong and Cao, Xiaochun},
  booktitle = 	 {ICML},
  pages = 	 {25750--25761},
  year = 	 {2025},
  volume = 	 {267},
  series = 	 {Proceedings of Machine Learning Research},
  month = 	 {13--19 Jul},
  publisher =    {PMLR},
  url = 	 {https://proceedings.mlr.press/v267/huang25ad.html}
}

@article{zhang2024holmes,
  title={{Holmes-VAD}: Towards unbiased and explainable video anomaly detection via multi-modal {LLM}},
  author={Zhang, Huaxin and Xu, Xiaohao and Wang, Xiang and Zuo, Jialong and Han, Chuchu and Huang, Xiaonan and Gao, Changxin and Wang, Yuehuan and Sang, Nong},
  journal={arXiv preprint arXiv:2406.12235},
  year={2024}
}

@article{zhu2025vau,
  title={{VAU-R1}: Advancing video anomaly understanding via reinforcement fine-tuning},
  author={Zhu, Liyun and Chen, Qixiang and Shen, Xi and Cun, Xiaodong},
  journal={arXiv preprint arXiv:2505.23504},
  year={2025}
}

@misc{wang2024neural,
      title={Neural Network Diffusion}, 
      author={Kai Wang and Dongwen Tang and Boya Zeng and Yida Yin and Zhaopan Xu and Yukun Zhou and Zelin Zang and Trevor Darrell and Zhuang Liu and Yang You},
      year={2024},
      eprint={2402.13144},
      archivePrefix={arXiv},
      primaryClass={cs.LG},
      url={https://arxiv.org/abs/2402.13144}, 
}

@inproceedings{
    schurholt2021self,
    title={Self-Supervised Representation Learning on Neural Network Weights for Model Characteristic Prediction},
    author={Konstantin Sch{\"u}rholt and Dimche Kostadinov and Damian Borth},
    booktitle={NeurIPS},
    year={2021},
    url={https://openreview.net/forum?id=F1D8buayXQT}
}

@misc{peebles2022learning,
      title={Learning to Learn with Generative Models of Neural Network Checkpoints}, 
      author={William Peebles and Ilija Radosavovic and Tim Brooks and Alexei A. Efros and Jitendra Malik},
      year={2022},
      eprint={2209.12892},
      archivePrefix={arXiv},
      primaryClass={cs.LG},
      url={https://arxiv.org/abs/2209.12892}, 
}

@inproceedings{pratt2023does,
  title={What does a platypus look like? generating customized prompts for zero-shot image classification},
  author={Pratt, Sarah and Covert, Ian and Liu, Rosanne and Farhadi, Ali},
  booktitle={ICCV},
  pages={15691--15701},
  year={2023}
}

@inproceedings{liu2023visual,
    author = {Liu, Haotian and Li, Chunyuan and Wu, Qingyang and Lee, Yong Jae},
    booktitle = {NeurIPS},
    pages = {34892--34916},
    publisher = {Curran Associates, Inc.},
    title = {Visual Instruction Tuning},
    url = {https://proceedings.neurips.cc/paper_files/paper/2023/file/6dcf277ea32ce3288914faf369fe6de0-Paper-Conference.pdf},
    volume = {36},
    year = {2023}
}

@misc{geng2026mle,
      title={{MLE-UVAD}: Minimal Latent Entropy Autoencoder for Fully Unsupervised Video Anomaly Detection}, 
      author={Yuang Geng and Junkai Zhou and Kang Yang and Pan He and Zhuoyang Zhou and Jose C. Principe and Joel Harley and Ivan Ruchkin},
      year={2026},
      eprint={2603.23868},
      archivePrefix={arXiv},
      primaryClass={cs.CV},
      url={https://arxiv.org/abs/2603.23868}, 
}

@inproceedings{
    ha2016hypernetworks,
    title={HyperNetworks},
    author={David Ha and Andrew M. Dai and Quoc V. Le},
    booktitle={ICLR},
    year={2017},
    url={https://openreview.net/forum?id=rkpACe1lx}
}

@misc{hywu2026,
      title={{HY-WU} (Part I): An Extensible Functional Neural Memory Framework and An Instantiation in Text-Guided Image Editing}, 
      author={{Tencent HY Team}},
      year={2026},
      eprint={2603.07236},
      archivePrefix={arXiv},
      primaryClass={cs.CV},
      url={https://arxiv.org/abs/2603.07236}, 
}

@inproceedings{
    hu2022lora,
    title={Lo{RA}: Low-Rank Adaptation of Large Language Models},
    author={Edward J Hu and yelong shen and Phillip Wallis and Zeyuan Allen-Zhu and Yuanzhi Li and Shean Wang and Lu Wang and Weizhu Chen},
    booktitle={ICLR},
    year={2022},
    url={https://openreview.net/forum?id=nZeVKeeFYf9}
}

@inproceedings{cai2025hiprobe,
author = {Cai, Zhaolin and Li, Fan and Zheng, Ziwei and Qin, Yanjun},
title = {{HiProbe-VAD}: Video Anomaly Detection via Hidden States Probing in Tuning-Free Multimodal {LLM}s},
year = {2025},
isbn = {9798400720352},
publisher = {Association for Computing Machinery},
address = {New York, NY, USA},
url = {https://doi.org/10.1145/3746027.3755575},
doi = {10.1145/3746027.3755575},
booktitle = {Proceedings of the 33rd ACM International Conference on Multimedia},
pages = {592–601},
numpages = {10},
location = {Dublin, Ireland},
series = {MM '25}
}

@InProceedings{ye2025vera,
    author    = {Ye, Muchao and Liu, Weiyang and He, Pan},
    title     = {{VERA}: Explainable Video Anomaly Detection via Verbalized Learning of Vision-Language Models},
    booktitle = {CVPR},
    month     = {June},
    year      = {2025},
    pages     = {8679-8688}
}

@inproceedings{acsintoae2022ubnormal,
  title={Ubnormal: New benchmark for supervised open-set video anomaly detection},
  author={Acsintoae, Andra and Florescu, Andrei and Georgescu, Mariana-Iuliana and Mare, Tudor and Sumedrea, Paul and Ionescu, Radu Tudor and Khan, Fahad Shahbaz and Shah, Mubarak},
  booktitle={CVPR},
  pages={20143--20153},
  year={2022}
}

@INPROCEEDINGS{hivau70k,
  author={Zhang, Huaxin and Xu, Xiaohao and Wang, Xiang and Zuo, Jialong and Huang, Xiaonan and Gao, Changxin and Zhang, Shanjun and Yu, Li and Sang, Nong},
  booktitle={CVPR}, 
  title={Holmes-{VAU}: Towards Long-term Video Anomaly Understanding at Any Granularity}, 
  year={2025},
  volume={},
  number={},
  pages={13843-13853},
  doi={10.1109/CVPR52734.2025.01292}}

@inproceedings{yang2024follow,
author = {Yang, Yuchen and Lee, Kwonjoon and Dariush, Behzad and Cao, Yinzhi and Lo, Shao-Yuan},
title = {Follow the Rules: Reasoning for Video Anomaly Detection with Large Language Models},
year = {2024},
isbn = {978-3-031-73003-0},
publisher = {Springer-Verlag},
address = {Berlin, Heidelberg},
url = {https://doi.org/10.1007/978-3-031-73004-7_18},
doi = {10.1007/978-3-031-73004-7_18},
booktitle = {ECCV},
pages = {304–322},
numpages = {19},
location = {Milan, Italy}
}

@inproceedings{zanella2024harnessing,
  title={Harnessing Large Language Models for Training-free Video Anomaly Detection},
  author={Zanella, Luca and Menapace, Willi and Mancini, Massimiliano and Wang, Yiming and Ricci, Elisa},
  booktitle={CVPR},
  pages={18527--18536},
  year={2024}
}

@misc{bai2025qwen2,
      title={{Qwen2.5-VL} Technical Report}, 
      author={Shuai Bai and Keqin Chen and Xuejing Liu and Jialin Wang and Wenbin Ge and Sibo Song and Kai Dang and Peng Wang and Shijie Wang and Jun Tang and Humen Zhong and Yuanzhi Zhu and Mingkun Yang and Zhaohai Li and Jianqiang Wan and Pengfei Wang and Wei Ding and Zheren Fu and Yiheng Xu and Jiabo Ye and Xi Zhang and Tianbao Xie and Zesen Cheng and Hang Zhang and Zhibo Yang and Haiyang Xu and Junyang Lin},
      year={2025},
      eprint={2502.13923},
      archivePrefix={arXiv},
      primaryClass={cs.CV},
      url={https://arxiv.org/abs/2502.13923}, 
}

@misc{internvl25,
      title={Expanding Performance Boundaries of Open-Source Multimodal Models with Model, Data, and Test-Time Scaling}, 
      author={Zhe Chen and Weiyun Wang and Yue Cao and Yangzhou Liu and Zhangwei Gao and Erfei Cui and Jinguo Zhu and Shenglong Ye and Hao Tian and Zhaoyang Liu and Lixin Gu and Xuehui Wang and Qingyun Li and Yiming Ren and Zixuan Chen and Jiapeng Luo and Jiahao Wang and Tan Jiang and Bo Wang and Conghui He and Botian Shi and Xingcheng Zhang and Han Lv and Yi Wang and Wenqi Shao and Pei Chu and Zhongying Tu and Tong He and Zhiyong Wu and Huipeng Deng and Jiaye Ge and Kai Chen and Kaipeng Zhang and Limin Wang and Min Dou and Lewei Lu and Xizhou Zhu and Tong Lu and Dahua Lin and Yu Qiao and Jifeng Dai and Wenhai Wang},
      year={2025},
      eprint={2412.05271},
      archivePrefix={arXiv},
      primaryClass={cs.CV},
      url={https://arxiv.org/abs/2412.05271}, 
}

@InProceedings{liu2024improved,
    author    = {Liu, Haotian and Li, Chunyuan and Li, Yuheng and Lee, Yong Jae},
    title     = {Improved Baselines with Visual Instruction Tuning},
    booktitle = {CVPR},
    month     = {June},
    year      = {2024},
    pages     = {26296-26306}
}

@INPROCEEDINGS {thakare2023dyannet,
author = { Thakare, Kamalakar Vijay and Raghuwanshi, Yash and Dogra, Debi Prosad and Choi, Heeseung and Kim, Ig-Jae },
booktitle = { WACV },
title = {{ DyAnNet: A Scene Dynamicity Guided Self-Trained Video Anomaly Detection Network }},
year = {2023},
volume = {},
ISSN = {},
pages = {5530-5539},
doi = {10.1109/WACV56688.2023.00550},
url = {https://doi.ieeecomputersociety.org/10.1109/WACV56688.2023.00550},
publisher = {IEEE Computer Society},
address = {Los Alamitos, CA, USA},
month =Jan}

@INPROCEEDINGS{zaheer2022generative,
  author={Zaheer, M. Zaigham and Mahmood, Arif and Khan, M. Haris and Segu, Mattia and Yu, Fisher and Lee, Seung-Ik},
  booktitle={CVPR}, 
  title={Generative Cooperative Learning for Unsupervised Video Anomaly Detection}, 
  year={2022},
  volume={},
  number={},
  pages={14724-14734},
  doi={10.1109/CVPR52688.2022.01433}}

@inproceedings{park2020learning,
  title={Learning Memory-guided Normality for Anomaly Detection},
  author={Park, Hyunjong and Noh, Jongyoun and Ham, Bumsub},
  booktitle={CVPR},
  pages={14372--14381},
  year={2020}
}

@INPROCEEDINGS{gong2019memorizing,
  author={Gong, Dong and Liu, Lingqiao and Le, Vuong and Saha, Budhaditya and Mansour, Moussa Reda and Venkatesh, Svetha and Van Den Hengel, Anton},
  booktitle={ICCV}, 
  title={Memorizing Normality to Detect Anomaly: Memory-Augmented Deep Autoencoder for Unsupervised Anomaly Detection}, 
  year={2019},
  volume={},
  number={},
  pages={1705-1714},
  doi={10.1109/ICCV.2019.00179}}

@INPROCEEDINGS {wu2024open,
author = { Wu, Peng and Zhou, Xuerong and Pang, Guansong and Sun, Yujia and Liu, Jing and Wang, Peng and Zhang, Yanning },
booktitle = { CVPR },
title = {{ Open-Vocabulary Video Anomaly Detection }},
year = {2024},
volume = {},
ISSN = {},
pages = {18297-18307},
doi = {10.1109/CVPR52733.2024.01732},
url = {https://doi.ieeecomputersociety.org/10.1109/CVPR52733.2024.01732},
publisher = {IEEE Computer Society},
address = {Los Alamitos, CA, USA},
month =Jun}

@inproceedings{zhou2023dual,
author = {Zhou, Hang and Yu, Junqing and Yang, Wei},
title = {Dual memory units with uncertainty regulation for weakly supervised video anomaly detection},
year = {2023},
isbn = {978-1-57735-880-0},
publisher = {AAAI Press},
url = {https://doi.org/10.1609/aaai.v37i3.25489},
doi = {10.1609/aaai.v37i3.25489},
booktitle = {AAAI},
articleno = {420},
numpages = {9},
series = {AAAI'23/IAAI'23/EAAI'23}
}

@inproceedings{wu2024vadclip,
author = {Wu, Peng and Zhou, Xuerong and Pang, Guansong and Zhou, Lingru and Yan, Qingsen and Wang, Peng and Zhang, Yanning},
title = {{VadCLIP}: adapting vision-language models for weakly supervised video anomaly detection},
year = {2024},
isbn = {978-1-57735-887-9},
publisher = {AAAI Press},
url = {https://doi.org/10.1609/aaai.v38i6.28423},
doi = {10.1609/aaai.v38i6.28423},
booktitle = {AAAI},
articleno = {675},
numpages = {9},
series = {AAAI'24/IAAI'24/EAAI'24}
}

@INPROCEEDINGS{yang2024text,
  author={Yang, Zhiwei and Liu, Jing and Wu, Peng},
  booktitle={CVPR}, 
  title={Text Prompt with Normality Guidance for Weakly Supervised Video Anomaly Detection}, 
  year={2024},
  volume={},
  number={},
  pages={18899-18908},
  doi={10.1109/CVPR52733.2024.01788}}

@INPROCEEDINGS{joo2023clip,
  author={Joo, Hyekang Kevin and Vo, Khoa and Yamazaki, Kashu and Le, Ngan},
  booktitle={2023 IEEE International Conference on Image Processing (ICIP)}, 
  title={{CLIP-TSA}: Clip-Assisted Temporal Self-Attention for Weakly-Supervised Video Anomaly Detection}, 
  year={2023},
  volume={},
  number={},
  pages={3230-3234},
  doi={10.1109/ICIP49359.2023.10222289}}

@inproceedings{chen2023mgfn,
author = {Chen, Yingxian and Liu, Zhengzhe and Zhang, Baoheng and Fok, Wilton and Qi, Xiaojuan and Wu, Yik-Chung},
title = {{MGFN}: magnitude-contrastive glance-and-focus network for weakly-supervised video anomaly detection},
year = {2023},
isbn = {978-1-57735-880-0},
publisher = {AAAI Press},
url = {https://doi.org/10.1609/aaai.v37i1.25112},
doi = {10.1609/aaai.v37i1.25112},
booktitle = {AAAI},
articleno = {43},
numpages = {9},
series = {AAAI'23/IAAI'23/EAAI'23}
}

@inproceedings{li2022self,
  title={Self-Training Multi-Sequence Learning with Transformer for Weakly Supervised Video Anomaly Detection},
  author={S. Li and Fang Liu and Licheng Jiao},
  booktitle={AAAI},
  year={2022},
  url={https://api.semanticscholar.org/CorpusID:248982052}
}

@INPROCEEDINGS {tian2021weakly,
author = { Tian, Yu and Pang, Guansong and Chen, Yuanhong and Singh, Rajvinder and Verjans, Johan W. and Carneiro, Gustavo },
booktitle = { ICCV },
title = {{ Weakly-supervised Video Anomaly Detection with Robust Temporal Feature Magnitude Learning }},
year = {2021},
volume = {},
ISSN = {},
pages = {4955-4966},
doi = {10.1109/ICCV48922.2021.00493},
url = {https://doi.ieeecomputersociety.org/10.1109/ICCV48922.2021.00493},
publisher = {IEEE Computer Society},
address = {Los Alamitos, CA, USA},
month =Oct}

@inproceedings{ssrl,
  title={Scale-aware spatio-temporal relation learning for video anomaly detection},
  author={Li, Guoqiu and Cai, Guanxiong and Zeng, Xingyu and Zhao, Rui},
  booktitle={ECCV},
  pages={333--350},
  year={2022},
  organization={Springer}
}

@InProceedings{Kim_2025_ICCV,
    author    = {Kim, Younggun and Abdelrahman, Ahmed S. and Abdel-Aty, Mohamed},
    title     = {{VRU-Accident}: A Vision-Language Benchmark for Video Question Answering and Dense Captioning for Accident Scene Understanding},
    booktitle = {ICCV},
    month     = {October},
    year      = {2025},
    pages     = {761-771}
}

@ARTICLE{9712446,
author={Yao, Yu and Wang, Xizi and Xu, Mingze and Pu, Zelin and Wang, Yuchen and Atkins, Ella and Crandall, David J.},
journal={ IEEE T-PAMI },
title={{ {DoTA}: Unsupervised Detection of Traffic Anomaly in Driving Videos }},
year={2023},
volume={45},
number={01},
ISSN={1939-3539},
pages={444-459},
doi={10.1109/TPAMI.2022.3150763},
url = {https://doi.ieeecomputersociety.org/10.1109/TPAMI.2022.3150763},
publisher={IEEE Computer Society},
address={Los Alamitos, CA, USA},
month=jan}

@misc{shao2024deepseekmathpushinglimitsmathematical,
  author = {Zhihong Shao and Peiyi Wang and Qihao Zhu and Runxin Xu and Junxiao Song and Mingchuan Zhang and Y.K. Li and Y. Wu and Daya Guo},
  title = {{DeepSeekMath}: Pushing the Limits of Mathematical Reasoning in Open Language Models},
  journal = {CoRR},
  volume = {abs/2402.03300},
  year = {2024},
  url = {https://arxiv.org/abs/2402.03300},
}

@article{shen2025vlm,
  title={{VLM-R1}: A stable and generalizable r1-style large vision-language model},
  author={Shen, Haozhan and Liu, Peng and Li, Jingcheng and Fang, Chunxin and Ma, Yibo and Liao, Jiajia and Shen, Qiaoli and Zhang, Zilun and Zhao, Kangjia and Zhang, Qianqian and others},
  journal={arXiv preprint arXiv:2504.07615},
  year={2025}
}

@inproceedings{Wu2020not,
title={Not only Look, but also Listen: Learning Multimodal Violence Detection under 
Weak Supervision},
author={Wu, Peng and Liu, Jing and Shi, Yujia and Sun, Yujia and Shao, Fangtao 
and Wu, Zhaoyang and Yang, Zhiwei},
booktitle={ECCV},
year={2020}
}

@InProceedings{Anderson2016SPICESP,
author="Anderson, Peter
and Fernando, Basura
and Johnson, Mark
and Gould, Stephen",
title="{SPICE}: Semantic Propositional Image Caption Evaluation",
booktitle="ECCV",
year="2016",
address="Cham",
pages="382--398",
isbn="978-3-319-46454-1"
}

@inproceedings{lavie-agarwal-2007-meteor,
    title = "{METEOR}: An Automatic Metric for {MT} Evaluation with High Levels of Correlation with Human Judgments",
    author = "Lavie, Alon  and
      Agarwal, Abhaya",
    booktitle = "Proceedings of the Second Workshop on Statistical Machine Translation",
    month = jun,
    year = "2007",
    address = "Prague, Czech Republic",
    publisher = "Association for Computational Linguistics",
    url = "https://aclanthology.org/W07-0734/",
    pages = "228--231"
}

@inproceedings{rei-etal-2020-comet,
    title = "{COMET}: A Neural Framework for {MT} Evaluation",
    author = "Rei, Ricardo  and
      Stewart, Craig  and
      Farinha, Ana C  and
      Lavie, Alon",
    booktitle = "Proceedings of the 2020 Conference on Empirical Methods in Natural Language Processing (EMNLP)",
    month = nov,
    year = "2020",
    address = "Online",
    publisher = "Association for Computational Linguistics",
    url = "https://aclanthology.org/2020.emnlp-main.213/",
    doi = "10.18653/v1/2020.emnlp-main.213",
    pages = "2685--2702"
}

@inproceedings{lin-2004-rouge,
    title = "{ROUGE}: A Package for Automatic Evaluation of Summaries",
    author = "Lin, Chin-Yew",
    booktitle = "Text Summarization Branches Out",
    month = jul,
    year = "2004",
    address = "Barcelona, Spain",
    publisher = "Association for Computational Linguistics",
    url = "https://aclanthology.org/W04-1013/",
    pages = "74--81"
}

@inproceedings{papineni-etal-2002-bleu,
    title = "{B}leu: a Method for Automatic Evaluation of Machine Translation",
    author = "Papineni, Kishore  and
      Roukos, Salim  and
      Ward, Todd  and
      Zhu, Wei-Jing",
    booktitle = "Proceedings of the 40th Annual Meeting of the Association for Computational Linguistics",
    month = jul,
    year = "2002",
    address = "Philadelphia, Pennsylvania, USA",
    publisher = "Association for Computational Linguistics",
    url = "https://aclanthology.org/P02-1040/",
    doi = "10.3115/1073083.1073135",
    pages = "311--318"
}

@inproceedings{Vedantam2014CIDErCI,
  title={{CIDEr}: Consensus-based image description evaluation},
  author={Ramakrishna Vedantam and C. Lawrence Zitnick and Devi Parikh},
  booktitle={CVPR},
  year={2014},
  pages={4566-4575},
  url={https://api.semanticscholar.org/CorpusID:9026666}
}

@inproceedings{10.1007/978-3-031-19778-9_42,
author = {Wu, Jhih-Ciang and Hsieh, He-Yen and Chen, Ding-Jie and Fuh, Chiou-Shann and Liu, Tyng-Luh},
title = {Self-supervised Sparse Representation for Video Anomaly Detection},
year = {2022},
isbn = {978-3-031-19777-2},
publisher = {Springer-Verlag},
address = {Berlin, Heidelberg},
url = {https://doi.org/10.1007/978-3-031-19778-9_42},
doi = {10.1007/978-3-031-19778-9_42},
booktitle = {ECCV},
pages = {729–745},
numpages = {17},
location = {Tel Aviv, Israel}
}

@misc{Lv2024VideoAD,
      title={Video Anomaly Detection and Explanation via Large Language Models}, 
      author={Hui Lv and Qianru Sun},
      year={2024},
      eprint={2401.05702},
      archivePrefix={arXiv},
      primaryClass={cs.CV},
      url={https://arxiv.org/abs/2401.05702}, 
}

@inproceedings{dada2000,
  title={{Dada-2000: Can driving accident be predicted by driver attention? Analyzed by a benchmark}},
  author={Fang, Jianwu and Yan, Dingxin and Qiao, Jiahuan and Xue, Jianru and Wang, He and Li, Sen},
  booktitle={2019 IEEE Intelligent Transportation Systems Conference (ITSC)},
  pages={4303--4309},
  year={2019},
  organization={IEEE}
}

@article{ROUSSEEUW198753,
title = {Silhouettes: A graphical aid to the interpretation and validation of cluster analysis},
journal = {Journal of Computational and Applied Mathematics},
volume = {20},
pages = {53-65},
year = {1987},
issn = {0377-0427},
doi = {https://doi.org/10.1016/0377-0427(87)90125-7},
url = {https://www.sciencedirect.com/science/article/pii/0377042787901257},
author = {Peter J. Rousseeuw}
}

@article{JMLR:v9:vandermaaten08a,
  author  = {Laurens van der Maaten and Geoffrey Hinton},
  title   = {{Visualizing Data using t-SNE}},
  journal = {Journal of Machine Learning Research},
  year    = {2008},
  volume  = {9},
  number  = {86},
  pages   = {2579--2605},
  url     = {http://jmlr.org/papers/v9/vandermaaten08a.html}
}

@inproceedings{jiang-etal-2024-scaling,
    title = "Scaling Sentence Embeddings with Large Language Models",
    author = "Jiang, Ting  and
      Huang, Shaohan  and
      Luan, Zhongzhi  and
      Wang, Deqing  and
      Zhuang, Fuzhen",
    editor = "Al-Onaizan, Yaser  and
      Bansal, Mohit  and
      Chen, Yun-Nung",
    booktitle = "Findings of the Association for Computational Linguistics: EMNLP 2024",
    month = nov,
    year = "2024",
    address = "Miami, Florida, USA",
    publisher = "Association for Computational Linguistics",
    url = "https://aclanthology.org/2024.findings-emnlp.181/",
    doi = "10.18653/v1/2024.findings-emnlp.181",
    pages = "3182--3196"
}

@misc{zhu2025internvl3exploringadvancedtraining,
      title={InternVL3: Exploring Advanced Training and Test-Time Recipes for Open-Source Multimodal Models}, 
      author={Jinguo Zhu and Weiyun Wang and Zhe Chen and Zhaoyang Liu and Shenglong Ye and Lixin Gu and Hao Tian and Yuchen Duan and Weijie Su and Jie Shao and Zhangwei Gao and Erfei Cui and Xuehui Wang and Yue Cao and Yangzhou Liu and Xingguang Wei and Hongjie Zhang and Haomin Wang and Weiye Xu and Hao Li and Jiahao Wang and Nianchen Deng and Songze Li and Yinan He and Tan Jiang and Jiapeng Luo and Yi Wang and Conghui He and Botian Shi and Xingcheng Zhang and Wenqi Shao and Junjun He and Yingtong Xiong and Wenwen Qu and Peng Sun and Penglong Jiao and Han Lv and Lijun Wu and Kaipeng Zhang and Huipeng Deng and Jiaye Ge and Kai Chen and Limin Wang and Min Dou and Lewei Lu and Xizhou Zhu and Tong Lu and Dahua Lin and Yu Qiao and Jifeng Dai and Wenhai Wang},
      year={2025},
      eprint={2504.10479},
      archivePrefix={arXiv},
      primaryClass={cs.CV},
      url={https://arxiv.org/abs/2504.10479}, 
}

\appendix

\section{Technical appendices and supplementary material}

\subsection{Traditional VAD}
Fully supervised VAD methods rely on fine-grained temporal annotations for precise localization, but suffer from high labeling cost and limited scalability~\cite{acsintoae2022ubnormal}. Weakly supervised approaches instead learn from video-level labels, typically via multiple instance learning (MIL) or contrastive objectives~\cite{sultani2018real, li2022self, tian2021weakly, chen2023mgfn, joo2023clip, yang2024text, wu2024vadclip, zhou2023dual, wu2024open}. Unsupervised methods further relax supervision by modeling normality and detecting deviations using reconstruction or prediction-based objectives~\cite{gong2019memorizing, park2020learning, zaheer2022generative, thakare2023dyannet, geng2026mle}.

\subsection{Implementation}

We instantiate a lightweight parameter generator that produces LoRA updates conditioned on visual inputs. While we share the overall design across models, we adapt dimensionality and layer mappings to match each backbone. For InternVL2-8B \cite{internvl25}, we generate parameters of rank $r=8$,
%(higher ranks resulted in out-of-memory issues)
 operating over all transformer layers and targeting both attention and feed-forward modules. We set the generator depth to half that of the backbone (16 layers for a 32-layer language model). COPRA uses an 87.37M-parameter generator while keeping the underlying vision-language backbone frozen.
%Generated LoRA parameters are injected into fused attention projections (QKV and output) and feed-forward layers (gate, up, and down projections). 
For Qwen2-VL-7B-Instruct \cite{bai2025qwen2}, we use an analogous configuration. % with adjusted dimensionality (4096 to 3584)
%, while keeping the same rank and scaling factors for fair comparison. 
%We apply generated LoRA parameters to attention (query, key, value, output) and MLP projections (gate, up, down). 
%Although different backbones %(e.g., Qwen2-VL vs. InternVL2) 
%differ in hidden dimensionality, %(e.g., 3584 vs. 4096)
Across different backbones,
we standardize the internal generator dimension (512).
%, with head counts adjusted accordingly (e.g., 4096/512 = 8 heads for InternVL2-8B), ensuring consistent parameter generation across architectures.

%Across all experiments, we use FlashAttention-2 for efficient attention computation. 
We adopt the default VLM-R1~\cite{shen2025vlm} training configuration, retaining all hyperparameters unless adjustments are required to meet memory constraints. Models are trained for one epoch over the full training dataset. We use a learning rate of $1\times10^{-6}$ with linear decay, gradient clipping with a maximum norm of 1, no gradient accumulation, a KL regularization weight of $\beta=0.1$, and an importance sampling ratio clipping coefficient of $\epsilon=0.2$. Optimization is performed using AdamW with $\beta_1=0.9$, $\beta_2=0.999$, and $\varepsilon=1\times10^{-8}$. All experiments are conducted on a single NVIDIA A100-SXM4 GPU (80GB). Training and evaluation take approximately three days on UCF-Crime and XD-Violence.
\subsection{Prompt}
\begin{lstlisting}[style=paperstyle, language=python, label=lst:system_prompt, caption={System Prompt for VAD Classification}, showstringspaces=false]
You are the model.
** Model Description: **
You are designed to do binary classification. The input is a sequence of video frames for identifying whether there is an anomaly in the video; you need to output the class label, i.e., an integer in the set {0, 1}. 0 represents normal video, and 1 represents abnormal video. Please answer the prompt questions.

** Prompt Questions: **
Answer the following questions based on what you see from the video frames and provide an explanation in one sentence. 
Based on the analysis above, please conclude your answer to 'Is there any anomaly in the video?' in 'Yes, there is an anomaly' or 'No, there is no anomaly'.

Please give your output strictly in the following format:
<think>[Add your reasoning here]</think>
<answer>[ONLY the integer class label; make necessary assumptions if needed]</answer>

Please ONLY reply according to this format, don't give me any other words.
\end{lstlisting}

\begin{table}
\centering
\tablestyle{6pt}{1.2}

\caption{
Cross-dataset generalization from UCF-Crime to XD-Violence using InternVL2-8B.
The model is trained on UCF-Crime and evaluated on the unseen XD-Violence dataset.
Compared with the frozen baseline, COPRA improves AUC at every inference stage.
}
\label{tab:internvl_generalization_breakdown}

% \resizebox{\linewidth}{!}{%
\begin{tabular}{l>{\columncolor{baselinecolor}}c>{\columncolor{nvidiagreen}}cc}
\multirow{2}{*}{Inference Stage}
& Frozen & \textbf{COPRA} & Improvement \\
& Baseline &  & ($\uparrow$) \\
\shline

Initial (Step 1)
& 71.78 & \textbf{75.90} & {\color{mygreen}{+4.12}} \\

Initial + Retrieval (Step 2)
& 89.53 & \textbf{90.20} & {\color{mygreen}{+0.67}} \\

Initial + Retrieval + Smoothing + Weighting (Step 3)
& 90.63 & \textbf{90.98} & {\color{mygreen}{+0.35}} \\

\end{tabular}%
% }

\end{table}

\subsection{Reward Design and GRPO Objective}

Our video-level rewards are:
\begin{align}
\text{Accuracy}(s_0,a_0,\dots,a_O) &= \begin{cases} 1 & \text{output} == \text{ground truth}\\
0 & \text{otherwise} \end{cases}\\
\text{Format}(s_0,a_0,\dots,a_O) &= \begin{cases} 1 & \text{output matches <think>...</think><answer>...</answer>}\\
0&\text{otherwise} \end{cases}
\end{align}
We deliberately use binary rewards for their simplicity and verifiability — a simple choice in RL settings to isolate the effectiveness of our approach. Our loss is given by: \cite{shao2024deepseekmathpushinglimitsmathematical}

% \begin{equation*}
%     \Delta_\theta = g_\varphi (V)
% \end{equation*}
\begin{equation*}
\resizebox{\textwidth}{!}{$
\displaystyle
L_{\text{GRPO}} =
-\mathbb{E}_{\{(s_0, a^g_0, \dots, a^g_O)\}_{g=1}^{G} \sim \pi_{\theta+\Delta_\theta}(\cdot \mid s_0)}
\left[
\frac{1}{O} \sum_{i=1}^{O}
\min\left(
\frac{\pi_{\theta+\Delta_\theta}(a^g_i \mid s_0, a^g_0, \dots, a^g_{i-1})}
{\pi_{\theta_{\text{old}}}(a^g_i \mid s_0, a^g_0, \dots, a^g_{i-1})}
\hat{A}^{g},
\operatorname{clip}\left(
\frac{\pi_{\theta+\Delta_\theta}(a^g_i \mid s_0, a^g_0, \dots, a^g_{i-1})}
{\pi_{\theta_{\text{old}}}(a^g_i \mid s_0, a^g_0, \dots, a^g_{i-1})},
1-\epsilon,
1+\epsilon
\right)
\hat{A}^{g}
\right)
\right]
$}
\end{equation*}
\begin{equation*}
    L_{\text{total}} = L_{GRPO} + \beta \mathrm{D}_{\text{KL}}(\pi_{\theta+\Delta_\theta} \mid \mid \pi_{\text{REF}})
\end{equation*}

Here, $s_0$ denotes the initial state consisting of the system prompt and sampled video frames, and $O$ is the number of output tokens per generation $g \in [1,G]$. The system prompt is provided in Listing \ref{lst:system_prompt}. $\pi_{REF}$ is the frozen language model, before adaptation. 

The advantage $\hat{A}^{g}$ is computed via group-wise normalization:
\begin{align*}
    R(s_0,a_0,\dots,a_O) &= \text{Accuracy}(s_0,a_0,\dots,a_O) + \text{Format}(s_0,a_0,\dots,a_O)\\
    \hat{A}^g &= \frac{R(s_0,a^g_0,\dots,a^g_O) - \text{mean}( \{ R(s_0,a^{g'}_0,\dots,a^{g'}_O) \}_{g'=1}^{G} )}{\text{std}( \{ R(s_0,a^{g'}_0,\dots,a^{g'}_O) \}_{g'=1}^{G} ) + 10^{-4}}
\end{align*}

Importantly, the $G$ sampled generations are all produced from a single adapted policy $\pi_\theta$ using temperature-based decoding, identical to standard token sampling in large language models. That is, diversity arises from stochastic decoding under fixed adapted parameters for a given video instance, rather than from generating each response with a different parameter adaptation for that same video. 
This is sufficient for meaningful advantage estimation because stochastic output sequences $\{a^g_0,\dots,a^g_O\}_{g=1}^G \sim \pi_{\theta}(\cdot | s_0)$ under the same adapted parameters produce differing predictions and rewards. The resulting reinforcement signal has a natural structure: when a majority of output sequences are correct, the generator receives net positive reinforcement for the parameters it produced for that instance; when all $G$ output sequences are uniformly correct, the normalized advantages collapse to zero and the generator receives no update; 
and when a majority are incorrect, the generator receives net negative reinforcement, discouraging the generator from producing similar parameters when conditioned on visually similar instances. This asymmetry ensures that the parameter generator is consistently pushed toward producing adapters that yield reliable correct predictions, without requiring explicit supervision over what the parameters themselves should look like.

%Importantly, this approach enables effective fine-tuning of the parameter generator without requiring expensive chain-of-thought (CoT) annotations. In contrast to supervised fine-tuning (SFT), which depends on curated intermediate reasoning traces or target outputs, our RL-based formulation leverages only task-level rewards, making it significantly more scalable in weakly supervised settings such as VAD.

\begin{table}
\centering
\tablestyle{8pt}{1.2}

\caption{
Inference efficiency breakdown for InternVL2-8B with COPRA.
We report wall-clock time and peak GPU memory for the main inference components: frozen baseline response generation, parameter generation, and adapted response generation.
All measurements are averaged over 30 scoring windows with a maximum response length of 256 tokens.
}
\label{tab:copra_inference_efficiency}

\begin{tabular}{lcc}
\multirow{2}{*}{Inference Component}
& Time & VRAM \\
& (second) & (MB) \\
\shline

Baseline Response Generation
& $1.05892 \pm 0.26664$ & $2315.36048 \pm 1.35751$ \\

Parameter Generation
& $0.11824 \pm 0.08367$ & $832.08968 \pm 1.45848$ \\

COPRA Response Generation
& $1.17716 \pm 0.33232$ & $3147.45016 \pm 1.99249$ \\

\end{tabular}

\end{table}

\begin{table}
\centering
\tablestyle{8pt}{1.2}

\caption{
UCF-Crime test AUC using InternVL3-8B \cite{zhu2025internvl3exploringadvancedtraining} after one epoch of training.
Compared with the frozen baseline, COPRA improves AUC at every inference stage and also exceeds the InternVL2-8B COPRA result on UCF-Crime, which achieves 87.14\% AUC.
}
\label{tab:internvl3_ucf_one_epoch_breakdown_ucf}

\begin{tabular}{l>{\columncolor{baselinecolor}}c>{\columncolor{nvidiagreen}}cc}
\multirow{2}{*}{Inference Stage}
& Frozen & \textbf{COPRA} & Improvement \\
& Baseline &  & ($\uparrow$) \\
\shline

Initial (Step 1)
& 74.77 & \textbf{76.24} & {\color{mygreen}{+1.47}} \\

Initial + Retrieval (Step 2)
& 84.90 & \textbf{85.84} & {\color{mygreen}{+0.94}} \\

Initial + Retrieval + Smoothing + Weighting (Step 3)
& 86.62 & \textbf{87.33} & {\color{mygreen}{+0.71}} \\

\end{tabular}

\end{table}

\begin{table}[!htb]
\centering
\tablestyle{8pt}{1.2}

\caption{
XD-Violence test AP using InternVL3-8B after one epoch of training.
Compared with the frozen baseline, COPRA improves AUC at every inference stage and also exceeds the InternVL2-8B COPRA result on UCF-Crime at Step 1, which achieves 67.24\% AP.
}
\label{tab:internvl3_ucf_one_epoch_breakdown_xd}

\begin{tabular}{l>{\columncolor{baselinecolor}}c>{\columncolor{nvidiagreen}}cc}
\multirow{2}{*}{Inference Stage}
& Frozen & \textbf{COPRA} & Improvement \\
& Baseline &  & ($\uparrow$) \\
\shline

Initial (Step 1)
& 67.08 & \textbf{69.82} & {\color{mygreen}{+2.74}} \\

Initial + Retrieval (Step 2)
& 72.86 & \textbf{74.81} & {\color{mygreen}{+1.95}} \\

Initial + Retrieval + Smoothing + Weighting (Step 3)
& 73.96 & \textbf{75.70} & {\color{mygreen}{+1.73}} \\

\end{tabular}

\end{table}

\subsection{Cross-Dataset Generalization Analysis for InternVL2-8B}

Although we report Average Precision (AP) as the primary metric in the main paper, following the standard evaluation protocol for video anomaly detection, we analyze AUC to assess whether COPRA improves threshold-independent ranking quality across datasets. Table~\ref{tab:internvl_generalization_breakdown} evaluates InternVL2-8B trained on UCF-Crime and tested directly on the unseen XD-Violence benchmark, showing that COPRA improves over the frozen baseline at every inference stage. These gains indicate that instance-conditioned parameter generation improves cross-domain transfer before retrieval-based refinement, and that COPRA preserves the advantage under the full inference pipeline.

\subsection{Inference Efficiency}
COPRA generates approximately one million LoRA parameters for an 8B-parameter backbone, which introduces additional inference overhead. To make this cost transparent, we report three separate timing components: the response time of the frozen InternVL2-8B baseline, the time required by the parameter generator to produce instance-conditioned LoRA parameters, and the response time of the augmented backbone after applying the generated parameters. We separate these components because parameter generation does not necessarily occur for every model response. Depending on the inference granularity, the generated parameters may be reused across an entire video, a long temporal chunk, or multiple 10-second segments, allowing us to amortize the parameter generation cost over an entire video. We report memory usage in Table ~\ref{tab:copra_inference_efficiency} using the same decomposition, including the baseline backbone, the parameter generator, and the augmented backbone, to provide a complete view of the computational cost introduced by instance-conditioned parameter generation.

\subsection{Scaling to InternVL3-8B}

We further evaluate COPRA with InternVL3-8B~\cite{zhu2025internvl3exploringadvancedtraining} after one epoch of training to test whether instance-conditioned adaptation remains effective for a stronger backbone. The InternVL3-8B parameter generator follows the same COPRA design as InternVL2-8B, including rank-$r=8$ LoRA updates, $\alpha=16$, an internal generator dimension of 512, and bfloat16 inference, but adapts its dimensionality and target-module mapping to the evolved InternVL3 architecture. Specifically, InternVL3-8B uses a 28-layer language backbone with hidden size 3584 and unfused attention projections, so COPRA targets the query, key, value, output, gate, up, and down projections separately. In contrast, InternVL2-8B uses a 32-layer backbone with hidden size 4096 and fused query-key-value attention, requiring different LoRA output shapes. Thus, the differences in generator configuration reflect architectural changes between InternVL2 and InternVL3 rather than changes to the COPRA method itself.

On UCF-Crime, COPRA improves over the frozen baseline at every inference stage, increasing AUC from 74.77 to 76.24 at Step 1 and from 86.62 to 87.33 after retrieval, smoothing, and weighting (Table~\ref{tab:internvl3_ucf_one_epoch_breakdown_ucf}). On XD-Violence, COPRA similarly improves AP across all stages, with the final post-processed score increasing from 73.96 to 75.70 (Table~\ref{tab:internvl3_ucf_one_epoch_breakdown_xd}). These results suggest that COPRA's gains are not specific to InternVL2-8B, but persist when scaling to InternVL3-8B.

\begin{figure}
    \centering
    \includegraphics[width=\linewidth]{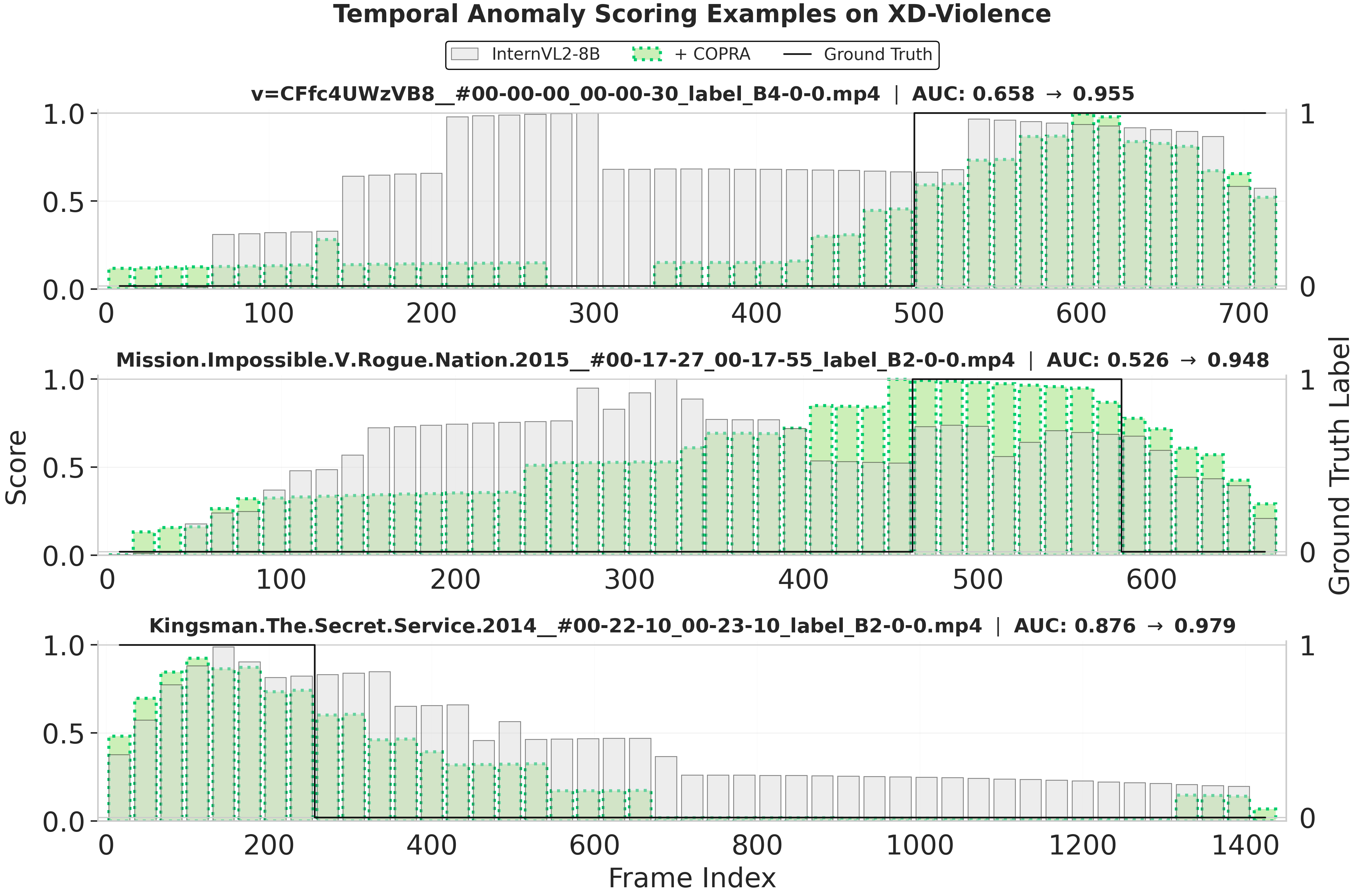}
    \caption{
    Qualitative temporal anomaly scoring example on XD-Violence.
    The grey bars show the InternVL2-8B baseline, the green bars show COPRA, and the black dashed signal indicates the ground-truth anomaly interval.
    Although the baseline assigns high anomaly scores to normal frames before the annotated event and drops during much of the abnormal interval, COPRA better aligns its peak response with the ground-truth abnormal segment.
    These examples illustrate how instance-conditioned parameter generation improves temporal anomaly localization.% under the same post-processing pipeline.
    }
    \label{fig:vad_prediction_example}
\end{figure}

\begin{figure}
    \centering
    \includegraphics[width=\linewidth]{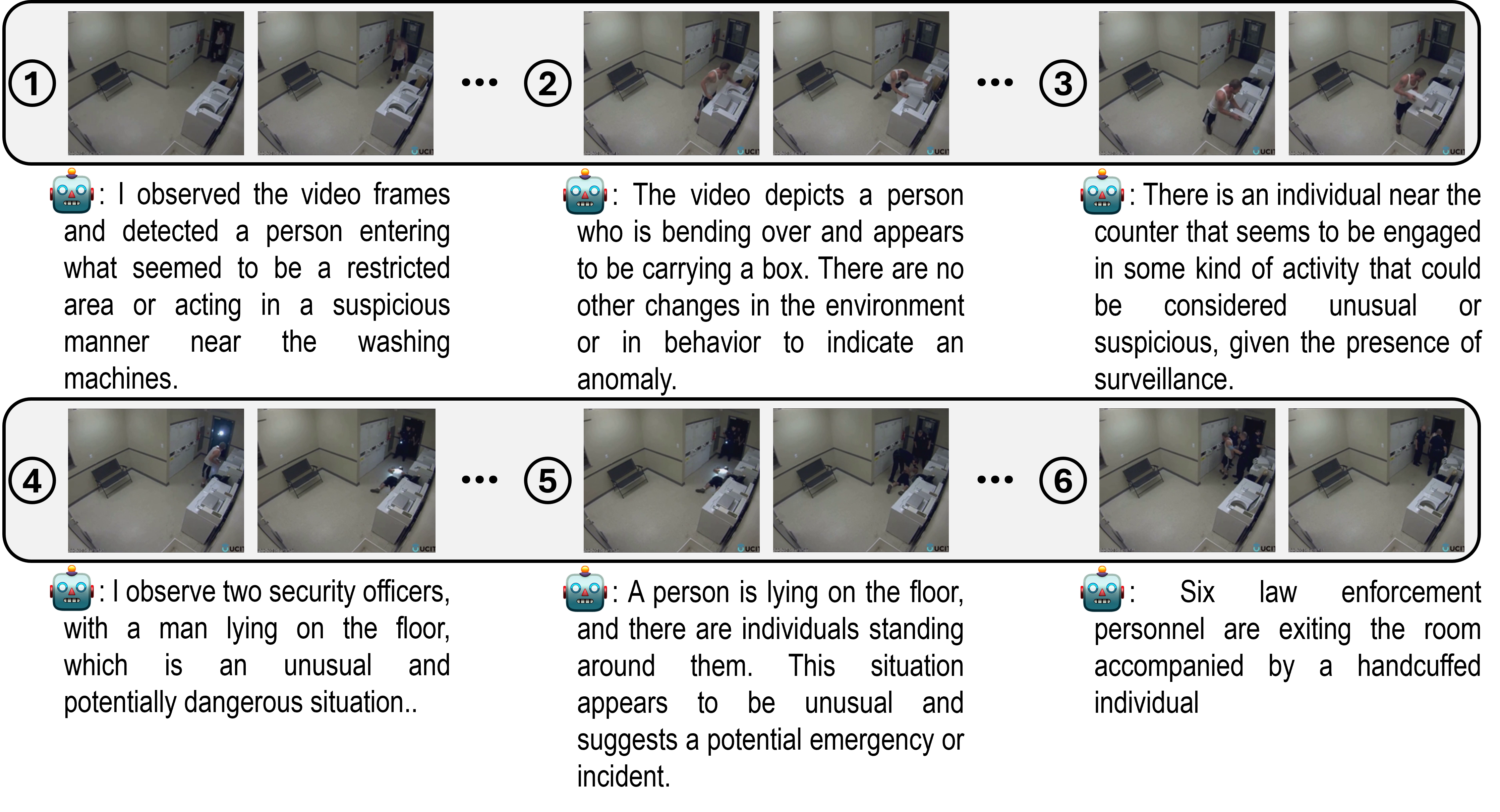}
    \caption{
        Illustration of COPRA's segment-level reasoning process on an arrest video from UCF-Crime.
        Each numbered group shows representative frames sampled from a temporal segment, where the visual context evolves from an initially empty room to suspicious human activity, physical confrontation, and the arrival of additional people.
        For each segment, the VLM produces a corresponding reasoning response, allowing the anomaly decision to be grounded in localized visual evidence and yielding a more reliable final anomaly score.
    }
    \label{fig:vad_reasoning_example}
\end{figure}

\subsection{Details of t-SNE
Embeddings and Silhouette Scores}

For the embedding visualization, we extract last-layer hidden states after the model processes the task prompt and visual input. As a lightweight diagnostic, we use the final input-token representation as a sequence-level embedding, yielding one embedding per input video--prompt pair. This choice is motivated by prior work that uses final-token hidden states as text representations~\cite{jiang-etal-2024-scaling}, although more specialized embedding methods may produce stronger representations. Under COPRA, the same input is processed with instance-conditioned parameters, which can shift the resulting summary embedding relative to the frozen baseline. We then project these high-dimensional embeddings into two dimensions using t-SNE~\cite{JMLR:v9:vandermaaten08a}, compute silhouette scores in the projected t-SNE space, and visualize the resulting representation space in Figure~\ref{fig:qualitative_analysis} (left).

For each video $V_i$, we compute the silhouette coefficient \cite{ROUSSEEUW198753} on the two-dimensional t-SNE embedding:
$s(V_i)=\frac{b(V_i)-a(V_i)}{\max\{a(V_i),b(V_i)\}}$, where $a(V_i)$ is the mean distance to videos from the same class and $b(V_i)$ is the mean distance to the nearest competing class. We report the average silhouette coefficient over all videos.

%%%%%%%%%%%%%%%%%%%%%%%%%%%%%%%%%%%%%%%%%%%%%%%%%%%%%%%%%%%%

% \newpage
% \input{checklist.tex}

\end{document}